\documentclass[10pt]{article}
\usepackage[preprint]{tmlr}
\usepackage{amsmath,amsfonts,bm}

\def\eqref#1{equation~\ref{#1}}
\def\1{\bm{1}}

\DeclareMathAlphabet{\mathsfit}{\encodingdefault}{\sfdefault}{m}{sl}
\SetMathAlphabet{\mathsfit}{bold}{\encodingdefault}{\sfdefault}{bx}{n}

\usepackage{hyperref}
\usepackage{url}
\usepackage{booktabs}
\usepackage{graphicx}
\usepackage{xcolor}
\usepackage{amsmath,amsthm}
\usepackage{float}
\usepackage{tikz}
\usetikzlibrary{positioning,arrows.meta,fit,backgrounds}
\hypersetup{colorlinks=true,linkcolor=red!50!black,citecolor=blue!50!black,urlcolor=blue!60!black}

\newcommand{\NGapEpisodes}{60}
\newcommand{\NContexts}{124{,}416}

\newcommand{\SelErrOutcome}{0.504}

\newcommand{\SelErrSuff}{0.208}

\newcommand{\SelErrNoiseLo}{0.043}

\newcommand{\SelErrCoarsePayload}{0.030}

\newcommand{\DecDisDropRFour}{0.1011}

\newcommand{\NCorpusPolicies}{14}
\newcommand{\NCorpusContexts}{5{,}256}
\newcommand{\NCorpusHeldout}{5}
\newcommand{\SlopeOpt}{1.81}
\newcommand{\SlopePess}{0.77}
\newcommand{\SlopeRatio}{2.3}
\newcommand{\RTwoOpt}{0.94}
\newcommand{\RTwoPess}{0.90}
\newcommand{\ControlTrue}{0.871}
\newcommand{\NLawCells}{13}
\newcommand{\NOpt}{6}
\newcommand{\NPess}{7}
\newcommand{\SlopeOptCI}{[1.67, 2.58]}
\newcommand{\SlopePessCI}{[0.26, 1.03]}
\newcommand{\SlopeRatioCI}{[1.7, 7.3]}
\newcommand{\PRatioGtOne}{1.00}
\newcommand{\LooMAE}{0.042}
\newcommand{\RTwoPooled}{0.92}

\newcommand{\RTwoOptStd}{0.72}
\newcommand{\RTwoPessStd}{0.78}
\newcommand{\SOnlyOptRTwo}{0.77}
\newcommand{\SOnlyPessRTwo}{0.88}

\newcommand{\SelErrFamOutcome}{0.453}

\newcommand{\SelErrFamSuff}{0.209}

\newcommand{\SelErrFamCoarse}{0.036}

\newcommand{\SelErrFamFamdrop}{0.066}

\newcommand{\SelErrFamJudge}{0.090}
\newcommand{\SignedFamJudge}{-0.067}
\newcommand{\TransBasePrec}{0.717}

\newcommand{\TransBaseFOne}{0.728}

\newcommand{\TransExactAPrec}{0.754}

\newcommand{\TransExactAFOne}{0.740}

\newcommand{\TransExactBFOne}{0.745}
\newcommand{\TransSuffPrec}{0.702}
\newcommand{\TransSuffRec}{0.965}
\newcommand{\TransSuffFOne}{0.779}

\newcommand{\TransSingleFOne}{0.750}

\newcommand{\CovFamStruct}{0.53}

\newcommand{\CovSingleStruct}{0.23}

\newcommand{\TransSuffRecCI}{[0.94, 0.99]}

\newcommand{\PermExactPrecP}{$p{=}0.30$}
\newcommand{\PermSuffRecP}{$p{<}0.001$}

\newcommand{\NTermEval}{96}
\newcommand{\TrueEndCoarseRun}{0.844}

\newcommand{\TrueEndCoarseRunB}{0.806}

\newcommand{\TrueEndExactRun}{0.838}

\newcommand{\TrueEndExactRunB}{0.905}

\newcommand{\TrueEndFamdropRun}{0.875}

\newcommand{\TrueEndFamdropRunB}{0.840}

\newcommand{\TrueEndJudgeRun}{0.865}

\newcommand{\EvalProxyStartJudgeRun}{0.884}

\newcommand{\TrueEndNoiseRunLo}{0.821}

\newcommand{\TrueEndNoiseRun}{0.678}

\newcommand{\TrueEndNoiseRunHi}{0.641}

\newcommand{\TrueEndOutcomeRun}{0.714}
\newcommand{\TrueEndSEOutcomeRun}{0.024}

\newcommand{\ProxyEndOutcomeRun}{1.000}

\newcommand{\SkipsOutcomeRun}{137}

\newcommand{\TrueEndRecallRun}{0.692}

\newcommand{\ProxyTermRecallRun}{0.974}

\newcommand{\CitedLastRecallRun}{7.4}
\newcommand{\CitedFirstRecallRun}{6.1}
\newcommand{\TrueEndSuffRun}{0.667}

\newcommand{\ProxyTermSuffRun}{0.940}
\newcommand{\GapTermSuffRun}{0.273}

\newcommand{\CitedLastSuffRun}{7.5}

\newcommand{\CitedLastSuffRunB}{8.6}

\newcommand{\EvalTrueStart}{0.731}
\newcommand{\NMlxRuns}{15}

\newcommand{\CoarseDamB}{+0.065}
\newcommand{\EvalTrueStartSE}{0.020}

\newcommand{\ControlGain}{0.140}
\newcommand{\ComputeTotalHours}{45}
\newcommand{\ComputeHaikuTotal}{9{,}076}
\newcommand{\ComputeJudgeCalls}{1{,}496}
\newcommand{\ComputeExtractCalls}{7{,}580}
\newcommand{\ComputeNFallback}{3}
\newcommand{\ComputeGenMinutes}{13}
\newcommand{\OracleChecked}{all 124,416 contexts with zero mismatches}

\newcommand{\RepMlxDam}{+0.218}
\newcommand{\RepMlxDamSE}{0.018}

\newcommand{\RepTQDam}{+0.084}

\newcommand{\RepTQMDam}{+0.083}

\newcommand{\RepTPCtrl}{0.672}
\newcommand{\RepTPDam}{+0.017}

\newcommand{\RepTPLCtrl}{0.764}
\newcommand{\RepTPLDam}{+0.087}
\newcommand{\RepTPLDamSE}{0.031}

\newcommand{\RepTQLDamA}{+0.061}
\newcommand{\RepTQLDamB}{+0.096}
\newcommand{\RepTQLDamMean}{+0.078}

\newcommand{\PhiThreeDam}{-0.011}
\newcommand{\PhiThreeDamSE}{0.037}
\newcommand{\PhiThreeInfl}{+1.38}
\newcommand{\RepMNoiseLoDam}{+0.024}

\newcommand{\RepMNoiseLoGap}{-0.038}
\newcommand{\RepMNoiseMidDam}{+0.018}

\newcommand{\RepMNoiseMidGap}{-0.121}
\newcommand{\RepMNoiseHiDam}{+0.071}

\newcommand{\RepMNoiseHiGap}{-0.185}
\newcommand{\RepSlopeRatioCI}{[1.0, 10.5]}
\newcommand{\RepPRatioGtOne}{0.98}
\newcommand{\RepSlopeOpt}{0.565}
\newcommand{\RepSlopePess}{0.175}
\newcommand{\RepSlopeRatio}{3.2}
\newcommand{\RepNOpt}{4}
\newcommand{\RepNPess}{4}
\newcommand{\RepRTwoOpt}{0.84}
\newcommand{\RepRTwoPess}{0.88}

\newcommand{\RepTQFourDam}{+0.034}
\newcommand{\RepTQFourDamSE}{0.027}

\newcommand{\RepTQFourGap}{+0.141}

\newcommand{\RepMFourFamDam}{-0.050}
\newcommand{\RepMFourCoarseDam}{-0.046}
\newcommand{\RepRecallTrue}{0.675}
\newcommand{\RepRecallDam}{+0.122}
\newcommand{\RepRecallDamSE}{0.026}
\newcommand{\RepRecallSigma}{4.7}
\newcommand{\RepRecallInfl}{+2.85}
\newcommand{\RepRecallCited}{7.03}

\newcommand{\RepRecallProxy}{0.944}
\newcommand{\RepRecallSel}{0.193}
\newcommand{\RepRecallSigned}{+0.052}
\newcommand{\RepRecallSkips}{33}
\newcommand{\RepTQFpCtrl}{0.826}
\newcommand{\RepTQFpDam}{+0.076}
\newcommand{\RepTQFpDamSE}{0.024}

\newcommand{\RepFpDelta}{-0.007}
\newcommand{\RepFpVerdict}{F1 not met and F2 met}
\newcommand{\RepMFamDam}{-0.006}

\newcommand{\RepMCoarseDam}{+0.010}

\newcommand{\RepMOutDam}{+0.019}
\newcommand{\RepMOutProxy}{0.999}
\newcommand{\RepMOutSkips}{113}
\newcommand{\RepGapMin}{+0.14}
\newcommand{\RepGapMax}{+0.31}
\newcommand{\RepTorchDamMin}{+0.017}
\newcommand{\RepTorchDamMax}{+0.096}
\newcommand{\RepNTorchCells}{10}

\newcommand{\RepNTorchSig}{7}

\newcommand{\CorpProbeMisses}{0}
\newcommand{\CorpProbeTotal}{896}
\newcommand{\CorpSelErrOutcome}{0.488}
\newcommand{\CorpSignedOutcome}{+0.249}
\newcommand{\CorpSelErrSuff}{0.165}
\newcommand{\CorpSignedSuff}{+0.196}

\newcommand{\CorpSignedNoiseMid}{-0.052}

\newcommand{\CorpSignedDrop}{-0.175}
\newcommand{\CorpCtrl}{0.785}
\newcommand{\CorpCtrlSE}{0.029}
\newcommand{\CorpCtrlCited}{2.62}
\newcommand{\CorpSuffDam}{+0.033}
\newcommand{\CorpSuffDamSE}{0.043}

\newcommand{\CorpSuffInfl}{+0.21}
\newcommand{\CorpOutcomeDam}{+0.026}

\newcommand{\CorpOutcomeProxy}{1.000}

\newcommand{\CorpNoiseMidDam}{+0.008}

\newcommand{\CorpDropDam}{+0.017}

\newcommand{\CorpCtrlSkips}{85}

\newcommand{\CorpOutcomeSkips}{149}
\newcommand{\CorpSuffSkips}{116}
\newcommand{\CorpBase}{0.759}
\newcommand{\CorpBaseSE}{0.033}
\newcommand{\CorpCtrlGain}{+0.026}
\newcommand{\CorpCtrlGainSE}{0.044}

\newcommand{\CorpScore}{C1 met. C2 sufficiency-most-damaging met, citation inflation above 0.5 not met, outcome saturation met, stale policy harmless met, noise between them not met. C3 met (control 0.785 against the untrained base 0.759). C4 met ($\beta_+ = 0.94$ against $\beta_- = 0.07$). C5 not met (step-1 group reward 0.889, 85 of 150 control steps skipped)}
\newcommand{\SeedBSuffDam}{+0.093}

\newcommand{\SeedBOutcomeDam}{+0.094}

\newcommand{\SeedCellsDone}{7}
\newcommand{\SeedCellsWithin}{7}
\newcommand{\SeedBCtrl}{0.812}

\newtheorem{definition}{Definition}
\newtheorem{proposition}{Proposition}

\title{Optimistic Verifiers Get Hacked:\\Reward Hacking Under an Exact Process Oracle}

\author{\name Arunkumar V \email arunkumarv1530@gmail.com \\
      \addr University College of Engineering, BIT Campus, Anna University, Tiruchirappalli, India
      \AND
      \name Manoranjan Gandhudi \\
      \addr Central University of Karnataka, Kalaburagi, India
      \AND
      \name Gangadharan G. R. \\
      \addr National Institute of Technology Tiruchirappalli, India
      \AND
      \name Arun Prakash \\
      \addr Jawaharlal Nehru University, New Delhi, India
      \AND
      \name S. Senthilkumar \\
      \addr University College of Engineering, BIT Campus, Anna University, Tiruchirappalli, India}

\def\month{11}
\def\year{2026}
\def\openreview{}

\begin{document}
\maketitle

\begin{abstract}
Reinforcement learning from verifiable rewards inherits whatever its verifier
gets wrong. In mathematics and code the verifier is exact. Outside them it is
a proxy, a rubric, a learned judge or a hand-written checker, and its distance
from the true objective is never measured, so a rising reward cannot be
distinguished from a hack. We build a setting where the true reward is exactly
computable at the process level. Executable authorization policies over an
enumerable space of $\NContexts$ contexts admit an oracle that lists every
minimal sufficient reason for every decision, which turns the causal fidelity
of a model's cited reasons into a computed quantity. We compile such policies
into exact process rewards, degrade the verifier along seven realistic axes
(outcome-only, sufficiency, recall-only, extraction noise, coarse taxonomy,
stale policy, LLM judge), and train against each while the exact verifier
scores every rollout. Four phenotypes appear, each with its own behavioural
signature. Optimistic verifiers are hacked by padding: the sufficiency proxy
terminates at \ProxyTermSuffRun{} while true faithfulness falls from
\EvalTrueStart{} to \TrueEndSuffRun{} on $n{=}\NTermEval$ held-out episodes.
Outcome-only verifiers saturate and stop teaching. Noise dilutes, and
ranking-preserving gaps do no harm. Which phenotype a run will produce is
predicted beforehand by a model-free audit of the verifier alone, its
selection error and gap sign, together with a short pilot of how often its
reward saturates. Across nine configurations spanning two frameworks, two
model vendors, two adapter sizes, three precisions, two initialisations and
two seeds, the audit quantity is invariant, the phenotypes recur, sufficiency
damage is positive in all \RepNTorchCells{} replication cells and beyond two
standard errors in \RepNTorchSig{}, and optimistic gaps stay the steeper
branch on the replication grid too (\RepSlopeRatio$\times$ against
\SlopeRatio$\times$, at an interval too wide to test the size). The damage
magnitude does not replicate, and five pre-registered attempts to explain it
are reported as they fell. We release the testbed, the training
instrumentation, and a second task family of \NCorpusPolicies{} real public
policies (Cedar, OPA, Gatekeeper) ported to executable oracles, on which the
audit reads the same way.
\end{abstract}

\section{Introduction}
\label{sec:intro}

Reinforcement learning with verifiable rewards (RLVR), where a program
checks each answer and returns the reward, has driven the recent
progress of reasoning models. In mathematics and code the verifier is exact.
Nearly everywhere else it is a
proxy, a rubric \citep{gunjal2025rubrics}, a reference-derived signal
\citep{rlvrr2026}, a learned judge or a hand-built checker. Reward hacking,
a policy raising its trained reward while the true objective
falls, is studied intensively
\citep{amodei2016concrete,skalse2022defining,gao2023scaling,pan2022effects,countdowncode2026,rewardhackinterventions2026,llmsgaming2026}.
Recent work argues that noisy verifiers may be tolerable
\citep{imperfectverifier2026}, and other work shows that systematic verifier
errors shape outcomes in ways an error rate does not predict
\citep{egashira2026delay}. All of this work reads hacking off downstream evaluations, probes,
held-out suites or anecdote, because the true reward is unavailable for the
individual rollout, the sampled output where the behaviour arises.

We make it available. Executable authorization policies, the allow/block
gates that govern tool use for deployed agents, Kubernetes admission and
cloud access control, are white-box functions over a categorical context
space \citep{cutler2024cedar}. A gate's decision tensor can be materialised, and an oracle can
compute every minimal sufficient reason (MSR) of every decision exactly
\citep{shih2018symbolic,ignatiev2019abduction}. Whether each reason a
model cites is causally relevant then reduces to a tensor slice. That is
process-level ground truth, a reward on why a decision was reached and not
only on whether it was right, which outcome verifiers for math and code
cannot supply. The exactness costs generality. Our results hold for
policy-gate explanation with 4B-parameter policy models and one
policy-gradient recipe, and \S\ref{sec:limits} says how far we have tested
that they travel.

We make four contributions.

\textbf{(1) Policy-as-code is a free verifier
(\S\ref{sec:setting}, \S\ref{sec:dial}).} We compile executable policy gates
into exact outcome and process rewards, and formalise seven degraded
verifiers that each corrupt exactly one stage of the reward pipeline. They
mirror how practitioners build verifiers, by checking only outcomes,
substituting a sufficiency or recall proxy, tolerating extraction noise,
coarsening the taxonomy, holding a stale policy, or delegating to a large
language model (LLM) as judge.

\textbf{(2) A model-free gap audit (\S\ref{sec:audit}).} Before any
training, we measure each degraded verifier's distance from truth on a
procedurally generated probe grid whose citation sets are exactly controlled.
The outcome-only verifier, the construction underlying most RLVR, cannot
rank a faithful explanation above an unfaithful one better than chance
(selection error $\SelErrOutcome$), the sufficiency proxy inverts
$\SelErrSuff$ of selections, and a stale policy missing one rule
mis-decides $\DecDisDropRFour$ of all $\NContexts$ contexts.

\textbf{(3) An exact anatomy of the gap, and what survives replication
(\S\ref{sec:anatomy}, \S\ref{sec:replication}).} We train a 4B policy
against each degraded verifier while the exact verifier scores every rollout,
so that the divergence between proxy and truth is a per-step observable.
Four phenotypes, distinct patterns of training behaviour, emerge. The
sufficiency proxy is hacked by padding, and carries the citation-inflation
signature we predicted in advance. Outcome-only rewards saturate within a few
steps and starve training from then on. Noise dilutes. Ranking-preserving
gaps, an LLM judge among them, are harmless. The audit's sign and selection
error predict which of the four a run will produce. The sufficiency proxy's
damage, the true reward it costs against an exact-reward control, is
positive in every replication run and significant in most, and a second,
structurally different optimistic verifier is hacked harder still. On the
original grid the damage magnitude is also a tight function of selection
error, with optimistic gaps $\SlopeRatio\times$ steeper than pessimistic ones
across four verifiers. The ordering and the sign asymmetry
replicate. The constants do not.

\textbf{(4) A real-policy corpus and a damage-export test
(\S\ref{sec:transfer}).} We release \NCorpusPolicies{} public policies
(Cedar, Open Policy Agent and Gatekeeper) ported to executable gates with exact oracles
($\NCorpusContexts$ contexts, \NCorpusHeldout{} held out). On them the hacked
model's padding exports intact while aggregate $F_1$ hides it, and the audit
reads as it did on the synthetic family. Exact-reward training transfers
without loss and without a significant gain, and corpus training is starved
by the ease of the task, which we report as a boundary.

The sections after related work take these contributions in order, then
give recommendations and limitations.

\section{Related Work}
\label{sec:related}

\paragraph{RLVR and the verifier problem.} The successes of RLVR in
mathematics and code \citep{shao2024deepseekmath} rest on verifiers that are
exact by construction. Extending it elsewhere has produced a family of proxy
verifiers, from rubric-based rewards \citep{gunjal2025rubrics} and
reference-derived signal chains \citep{rlvrr2026} to empirical arguments that
noisy verifiers are tolerable up to double-digit error rates
\citep{imperfectverifier2026}. \citet{fuzzingverifiers2026} fuzz verifiers for exploitable bugs before
training, and, closest to us, \citet{egashira2026delay}
inject systematic verification errors on arithmetic and find that outcomes
range from delay to collapse in a way the overall error rate does not
predict. \citet{mitsuhashi2026compute} inject random false positives and
false negatives on GSM8K and find false negatives the more damaging. All of
these must treat the verifier as given, because in their domains no
independent measurement of its distance from truth exists. Here the
verifier's gap is itself the measured quantity. That answers the negative
result of \citeauthor{egashira2026delay} with a positive one, since damage is
predicted by selection error weighted by update flow (\S\ref{sec:law}), and
accounts for the sign asymmetry: pessimistic noise damages in both studies,
while an optimistic verifier damages more per unit of corruption
(\S\ref{sec:law}). The noise-tolerance claims pass our test only
conditionally, on reward granularity and on optimizer robustness
(\S\ref{sec:anatomy}).

\paragraph{Reward hacking.} \citet{skalse2022defining} formalise hacking,
\citet{gao2023scaling} measure overoptimisation against a synthetic learned
gold reward model, Countdown-Code \citep{countdowncode2026} and
\citet{rewardhackinterventions2026} observe hacking in outcome-level
environments, and \citet{mahmoud2026rubric} separate verifier failure from
rubric-design failure with a judge panel as reference. Two
recent studies audit a verifier before training. \citet{zhang2026leaky} show
that a static leakiness audit of code test suites predicts the rewarded
false-positive mass, and \citet{rajan2026auditing} find that 28.5\% of
SWE-bench Verified tasks accept an incorrect patch. Both measure how much wrong output a verifier accepts. Neither can measure what a policy learned
from it, because the gold signal is again a test suite or a human. What we
add is the limit case: an exact process-level true reward, degraded
semantically rather than randomly, so that damage can be attributed to
a named property of the verifier. The starvation phenotype we
isolate is the pathological extreme of the uniform-group problem that
motivates dynamic sampling in trainers of the group relative policy
optimisation (GRPO) family \citep{yu2025dapo}.

\paragraph{Process rewards and faithfulness.} Process reward models
\citep{lightman2024verify} are learned and inherit judge noise, and step-level faithfulness rewards
\citep{faithrl2026} rely on evidence-support judges. \citet{yuan2026vpr}
convert symbolic oracles into dense process rewards and note the
dependence on oracle quality without studying it. We study exactly that
dependence, with an oracle that is perfect and degraded by construction. \citet{ferreira2026truthful} add a causal-attribution signal to the
reward model after showing that preference optimisation erodes explanation
faithfulness, and \citet{hase2026cst} reward chains of thought that let a
simulator predict counterfactual outputs. Both measure faithfulness against
the model's own behaviour, whereas our oracle scores it against the decision
function's true reasons. Formal explainable AI (XAI) provides exact attribution for white-box
functions through abductive minimal sufficient reasons
\citep{shih2018symbolic,ignatiev2019abduction}, and GlassGate
\citep{glassgate2026} used it to show that frontier models' free-form
decision explanations are causally imprecise regardless of scale. We turn
that oracle from an evaluation device into a training-time instrument.

\paragraph{Agent-safety oracles.} Vera \citep{vera2026} verifies whether a harmful
outcome occurred but measures neither attribution nor verifier coverage,
Who-Tests-the-Testers \citep{wtt2026} audits benchmark coverage without an
oracle, and AgentDojo \citep{agentdojo2024}, MCPTox \citep{mcptox2026} and
Agent-SafetyBench \citep{agentsafetybench2024} test agents against curated
attacks. We study the verifier
that trains and audits the agent.

\section{Setting: Policies with an Exact Process Reward}
\label{sec:setting}

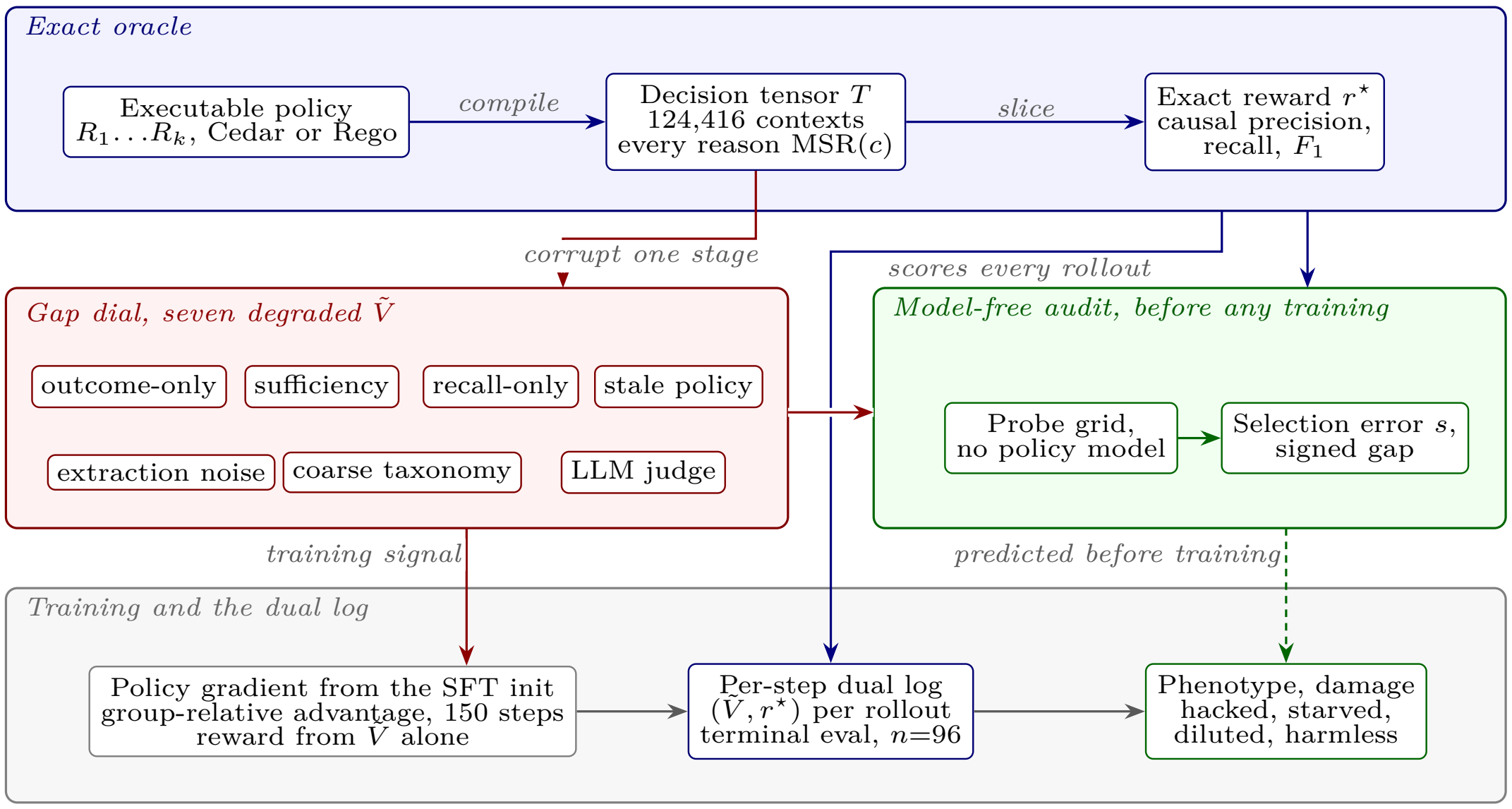
\begin{figure}[t]
\centering
\resizebox{0.86\textwidth}{!}{% Figure 1: the verifier-gap instrument.
% Blue = exact truth, vermillion = degraded proxy, green = what the instrument outputs.
% Grouped-panel layout. Every connector is routed orthogonally; no diagonals.
% Coordinates are explicit (cm) so the routing stays stable when content is edited.
% Kept narrow on purpose: the picture is \resizebox'd, so extra natural width
% shrinks every label. Measure before widening.
\begin{tikzpicture}[
  font=\small,
  x=1cm, y=1cm,
  ttl/.style={font=\scriptsize\itshape, inner sep=0pt, anchor=north west},
  bx/.style={draw, rounded corners=2pt, fill=white, align=center,
             inner sep=3pt, line width=0.45pt, font=\scriptsize},
  btruth/.style={bx, draw=blue!45!black},
  bout/.style={bx, draw=green!40!black},
  bneut/.style={bx, draw=black!50},
  chip/.style={draw, rounded corners=2pt, fill=white, draw=red!50!black,
               align=center, inner sep=2.4pt, line width=0.45pt, font=\scriptsize},
  arr/.style={-{Stealth[length=1.9mm]}, line width=0.6pt, black!65},
  tarr/.style={arr, draw=blue!50!black},
  parr/.style={arr, draw=red!55!black},
  oarr/.style={arr, draw=green!40!black},
  garr/.style={arr, draw=green!40!black, dashed, dash pattern=on 2pt off 1.6pt},
  lab/.style={font=\scriptsize\itshape, fill=white, inner sep=1.2pt},
  labA/.style={lab, fill=blue!5},   % sits inside panel A, must match its fill
  hop/.style={preaction={draw=white, line width=2.4pt, -}},
]

% =====================  PANEL A : the exact oracle  =====================
\fill[blue!5, rounded corners=3pt] (0,5.62) rectangle (14,7.52);
\draw[blue!45!black, rounded corners=3pt, line width=0.6pt] (0,5.62) rectangle (14,7.52);
\node[ttl, blue!45!black] at (0.16,7.44) {Exact oracle};

\node[btruth] (policy) at (2.15,6.45)
  {Executable policy\\[-1.5pt]$R_1{\dots}R_k$, Cedar or Rego};
\node[btruth] (oracle) at (7.00,6.45)
  {Decision tensor $T$\\[-1.5pt]$\NContexts$ contexts\\[-1.5pt]every reason $\mathrm{MSR}(c)$};
\node[btruth] (exact)  at (11.75,6.45)
  {Exact reward $r^\star$\\[-1.5pt]causal precision,\\[-1.5pt]recall, $F_1$};

\draw[tarr] (policy) -- node[labA,above] {compile} (oracle);
\draw[tarr] (oracle) -- node[labA,above] {slice}   (exact);

% =====================  PANEL B : the gap dial  =========================
\fill[red!5, rounded corners=3pt] (0,2.66) rectangle (7.30,4.90);
\draw[red!50!black, rounded corners=3pt, line width=0.6pt] (0,2.66) rectangle (7.30,4.90);
\node[ttl, red!55!black] at (0.16,4.82) {Gap dial, seven degraded $\tilde V$};

\node[chip] (c1) at (1.15,3.98) {outcome-only};
\node[chip] (c2) at (2.95,3.98) {sufficiency};
\node[chip] (c3) at (4.62,3.98) {recall-only};
\node[chip] (c4) at (6.28,3.98) {stale policy};
\node[chip] (c5) at (1.45,3.18) {extraction noise};
\node[chip] (c6) at (3.70,3.18) {coarse taxonomy};
\node[chip] (c7) at (5.95,3.18) {LLM judge};

% =====================  PANEL C : the model-free audit  =================
\fill[green!6, rounded corners=3pt] (8.10,2.66) rectangle (14,4.90);
\draw[green!40!black, rounded corners=3pt, line width=0.6pt] (8.10,2.66) rectangle (14,4.90);
\node[ttl, green!35!black] at (8.26,4.82) {Model-free audit, before any training};

\node[bout] (probe)  at (9.85,3.50) {Probe grid,\\[-1.5pt]no policy model};
\node[bout] (sigerr) at (12.50,3.50) {Selection error $s$,\\[-1.5pt]signed gap};
\draw[oarr] (probe) -- (sigerr);

% =====================  PANEL D : training and the dual log  ============
\fill[black!3, rounded corners=3pt] (0,0.10) rectangle (14,2.10);
\draw[black!50, rounded corners=3pt, line width=0.6pt] (0,0.10) rectangle (14,2.10);
\node[ttl, black!60] at (0.16,2.02) {Training and the dual log};

\node[bneut] (train) at (3.05,0.95)
  {Policy gradient from the SFT init\\[-1.5pt]group-relative advantage, 150 steps\\[-1.5pt]reward from $\tilde V$ alone};
\node[btruth] (log) at (7.70,0.95)
  {Per-step dual log\\[-1.5pt]$(\tilde V, r^\star)$ per rollout\\[-1.5pt]terminal eval, $n{=}\NTermEval$};
\node[bout] (pheno) at (11.95,0.95)
  {Phenotype, damage\\[-1.5pt]hacked, starved,\\[-1.5pt]diluted, harmless};

\draw[arr] (train) -- (log);
\draw[arr] (log)   -- (pheno);

% =====================  cross-panel routing (orthogonal only)  ==========
% oracle down into the dial
\draw[parr] (oracle.south) -- (7.00,5.36)
      -- node[lab, below, pos=0.60, inner sep=1.4pt] {corrupt one stage} (5.20,5.36)
      -- (5.20,4.90);
% exact reward down into the audit
\draw[tarr] (12.15,5.62) -- (12.15,4.90);
% exact reward down the panel gap into the dual log
\draw[tarr] (11.35,5.62) -- (11.35,5.24)
      -- node[lab, below, pos=0.52, inner sep=1.4pt] {scores every rollout} (7.70,5.24)
      -- (7.70,2.10) -- (7.70,0 |- log.north);
% dial across into the audit, hopping the vertical above
\draw[parr,hop] (7.30,3.74) -- (8.10,3.74);
% dial down into training
\draw[parr] (4.30,2.66) -- node[lab,left,pos=0.45] {training signal} (4.30,2.10)
      -- (4.30,0 |- train.north);
% audit down into the phenotype
\draw[garr] (11.95,2.66) -- node[lab,left,pos=0.45] {predicted before training} (11.95,2.10)
      -- (11.95,0 |- pheno.north);

\end{tikzpicture}}
\caption{\textbf{The verifier-gap instrument.} An executable policy gate is
compiled into an exact oracle (decision tensor and minimal sufficient
reasons), which defines the exact process reward $r^\star$. The gap dial
corrupts one stage of that pipeline to produce a degraded verifier
$\tilde V$. A model-free audit scores $\tilde V$ against $r^\star$ on a probe
grid before any training and yields a selection error and a sign. Training
sees only $\tilde V$, while $r^\star$ scores every rollout in parallel, so
the divergence between the two is logged per step. The dashed arrow is the
paper's claim, that the audit predicts the phenotype the run will produce.
The loop was replicated over MLX 4-bit on one laptop and PyTorch fp16
and NF4 on one T4 GPU, Qwen3-4B and Phi-3.5-mini, two adapters, two SFT lengths,
fp16 and fp32 adapters, two seeds, and a second task family of
\NCorpusPolicies{} real public policies where the audit transfers and
training is starved.}
\label{fig:arch}
\end{figure}

This section defines the policy gates, their exact process reward and the
episodes every experiment runs on; Fig.~\ref{fig:arch} shows the pipeline.

\paragraph{Policy gates.} A gate $G$ maps a context
$c \in \mathcal{X} = \prod_i V_i$ (ten categorical features covering tool,
scope, target domain, payload flags, auth level, role, intent and prior
violations) to a decision $G(c) \in \{\textsc{allow}, \textsc{block}\}$ by
first-match evaluation of an ordered rule list. Because
$|\mathcal{X}| = \NContexts$, $G$ materialises as a decision tensor $T$.

\begin{definition}[Sufficiency, MSR]
A feature set $S \subseteq [d]$ is \emph{sufficient} at $c$ iff every
completion of $c$'s values on $S$ preserves $G$'s decision. A \emph{minimal
sufficient reason} is a subset-minimal sufficient set, and $\mathrm{MSR}(c)$
is their antichain.
\end{definition}

\begin{definition}[Exact process reward]
For an explanation $t$ of decision $G(c)$ with cited feature set
$E(t) \subseteq [d]$ under a deterministic extractor $E$, causal precision
and recall are $P = |E(t) \cap R(c)|/|E(t)|$ with
$R(c) = \bigcup \mathrm{MSR}(c)$, and
$R = \max_{m \in \mathrm{MSR}(c)} |E(t) \cap m|/|m|$. The reward is
$r^\star(c,t) = F_1(P,R) - \lambda_b \mathbf{1}[\mathrm{len} > L] -
\lambda_d \mathbf{1}[\text{wrong decision word}]$.
\end{definition}

\begin{proposition}[Exactness]
For enumerable $\mathcal{X}$, sufficiency, $\mathrm{MSR}(c)$ and hence
$r^\star$ are computed exactly in $O(2^d)$ tensor slices per context. For
non-enumerable rule gates the same queries are decided exactly by abduction
over rule-relevant features (Appendix~\ref{app:oracle}).
\end{proposition}

\paragraph{Gate family and episodes.} Training spans a procedurally sampled
family of gates over the same schema, rejection-sampled to a block-rate band
of 0.35 to 0.65 with non-degenerate per-rule mass. There are 12 training gates and 4 held-out gates, never trained on, with 24
stratified episodes each. An episode is one context of one gate, with its
full oracle annotation, against which every reward in this paper is
computed. Real ported policies use the non-enumerating logical oracle, which agrees
with the tensor oracle on \OracleChecked{} (Appendix~\ref{app:oracle}).
\section{Seven Degraded Verifiers}
\label{sec:dial}

Each degraded verifier $\tilde{V}$ corrupts exactly one stage of the reward
pipeline just defined. The exact
verifier stays computable alongside, so
$\Delta(c,t) = \tilde{V}(c,t) - r^\star(c,t)$ is always observable.

\textbf{Outcome-only} $\tilde{V}_{\mathrm{out}}$ rewards the decision
word and ignores cited reasons, the default construction in RLVR.
\textbf{Sufficiency} $\tilde{V}_{\mathrm{suff}}$ rewards the graded
sufficiency of the cited set, the standard proxy in the
explanation-faithfulness literature \citep{deyoung2020eraser,lei2016rationalizing}. \textbf{Recall-only}
$\tilde{V}_{\mathrm{rec}}$ rewards causal recall with no precision term. It
is optimistic by a different mechanism, and tests whether the hack belongs to
one proxy or to optimism itself. \textbf{Extraction noise} $\tilde{V}_{\varepsilon}$ flips each
feature's cited status with probability $\varepsilon$, the random-noise
model of \citet{imperfectverifier2026}. \textbf{Coarse taxonomy}
$\tilde{V}_{\mathrm{coarse}}$ cannot distinguish features within a group,
such as the three payload flags, so credit flows to group-mates.
\textbf{Stale policy} $\tilde{V}_{\mathrm{drop}(r)}$ is missing rule $r$,
so its decisions and MSRs are computed on the wrong gate. \textbf{LLM
judge} $\tilde{V}_{\mathrm{judge}}$ replaces the deterministic extractor
with an LLM, the rubric-reward pattern. Formal definitions and the
invariants each verifier is tested against are in Appendix~\ref{app:dial}.

\section{Auditing a Verifier Before Training}
\label{sec:audit}

Before touching a policy model, we measure each verifier's distance from
truth on a procedurally generated probe grid. For stratified contexts
($n=\NGapEpisodes$), probe texts with exactly controlled citation sets span
the truth spectrum, from faithful and partial to padded, bare,
wrong-decision and distractor-only (Appendix~\ref{app:prompts}). We report the \emph{reward gap}
$\mathbb{E}|\tilde{V} - r^\star|$ and the \emph{selection error}, the
probability over probe pairs with different true rewards that the proxy
prefers the worse text, with ties counting one half. Selection error is the
distortion that drives best-of-$N$ selection and the sign of an RL advantage.
We also record the \emph{signed gap} $\mathbb{E}[\tilde V - r^\star]$: a
verifier is \emph{optimistic} where it is positive, scoring explanations
above their true reward, and \emph{pessimistic} where it is negative.

\begin{table}[t]
\centering
\caption{\textbf{Model-free gap audit.} Selection error is the probability
that the degraded verifier prefers the text with lower true reward. The
outcome-only verifier is a coin flip and the sufficiency proxy inverts one
selection in five. Stale-policy decision disagreement is exact over all
$\NContexts$ contexts, shown here for the least and most load-bearing rules (smallest and
largest decision disagreement)
(all eight in Table~\ref{tab:audit-full}, Appendix~\ref{app:dial}).}
\label{tab:audit}
% AUTO-GENERATED by experiments/make_macros.py -- do not edit.
\begin{tabular}{llccc}
\toprule
Verifier & Gap type & Reward gap & Selection error & Decision disagr. \\
\midrule
Exact (control) & n/a & 0.000 & 0.000 & n/a \\
Outcome-only & metric & 0.629 & 0.504 & n/a \\
Sufficiency proxy & metric & 0.239 & 0.208 & n/a \\
Recall-only proxy & metric & 0.090 & 0.132 & n/a \\
Extraction noise $\varepsilon{=}.05$ & extraction & 0.051 & 0.043 & n/a \\
Extraction noise $\varepsilon{=}.15$ & extraction & 0.156 & 0.143 & n/a \\
Extraction noise $\varepsilon{=}.30$ & extraction & 0.271 & 0.308 & n/a \\
Coarse taxonomy (payload) & granularity & 0.034 & 0.030 & n/a \\
Coarse taxonomy (session) & granularity & 0.076 & 0.048 & n/a \\
Stale policy (drop R4) & policy model & 0.057 & 0.042 & 0.1011 \\
Stale policy (drop R7) & policy model & 0.031 & 0.023 & 0.0229 \\
\bottomrule
\end{tabular}

\end{table}

Table~\ref{tab:audit} gives three findings. The outcome-only verifier has
selection error $\SelErrOutcome$, because faithful and unfaithful
explanations both carry the right decision word. The sufficiency proxy has
selection error $\SelErrSuff$, because padding is invisible to it, so it
prefers cite-everything texts over precise ones. And semantic gaps are not
interchangeable with random noise. Extraction noise at $\varepsilon = 0.05$
costs only $\SelErrNoiseLo$ selection error, while dropping one load-bearing
rule (R4) silently changes $\DecDisDropRFour$ of all decisions, a gap
invisible to any audit that lacks the true gate.

Three terms recur in the training results that follow. A \emph{cell} is one training run at one seed, a \emph{grid} is the
cells sharing a framework and recipe, and \emph{damage} is the exact
control's terminal true reward minus the cell's, so positive damage is
faithfulness the degraded verifier cost the policy.

\section{Training Against Degraded Verifiers}
\label{sec:anatomy}

\begin{figure}[t]
\centering
\includegraphics[width=\textwidth,height=0.78\textheight,keepaspectratio]{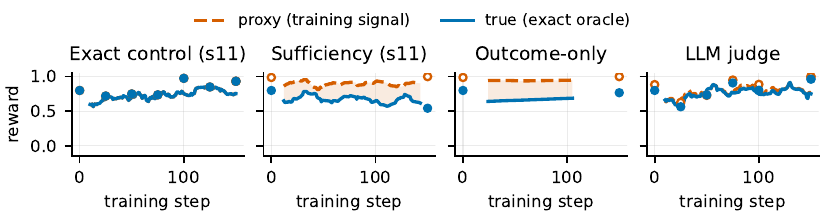}
\caption{\textbf{Hacking curves.} Rolling on-policy reward under the training
proxy (dashed, vermillion) and the exact oracle (solid, blue), with circles
for the 8-episode in-training evals and the shaded band for the observed
gap. The
control tracks truth by construction, the sufficiency proxy diverges from it
while reporting near-perfect reward, outcome-only saturates at once, and the
judge stays level with the control. All \NMlxRuns{} runs are in
Fig.~\ref{fig:curves_all}.}
\label{fig:curves}
\end{figure}

We train the same policy, initialised by supervised fine-tuning (SFT),
against each degraded verifier for 150 steps while the exact verifier scores every rollout
(Fig.~\ref{fig:curves}, Table~\ref{tab:training}). Three pathologies
emerge, a fourth gap type does no harm, and the proxy sees none of it.
The section closes by linking each phenotype to the audit.

\paragraph{The exact reward is worth training on.} Prompted
explanations sit at causal precision of about 0.50 to 0.54 from a 14B model
up to frontier models \citep{glassgate2026,turpin2023language,lanham2023measuring},
so content selection is a problem of objective, not capacity. From an SFT
initialisation scoring \EvalTrueStart{} $\pm$ \EvalTrueStartSE{} on held-out gates, 150
steps of policy gradient against the exact process reward reach
\TrueEndExactRun{} and \TrueEndExactRunB{} (seeds 11 and 12, mean
\ControlTrue), a \ControlGain{} gain on gates never seen in training. Every endpoint in this paper
is scored on the full held-out set ($n = \NTermEval$). The trainer is a
transparent REINFORCE loop \citep{williams1992reinforce} with group-relative advantage (the GRPO core at
on-policy single-step), gradient clipping and per-completion accumulation
(Appendix~\ref{app:training}).
An earlier unclipped configuration was seed-unstable and an off-the-shelf
GRPO library produced silent no-op checkpoints (Appendix~\ref{app:grpo}).

\begin{table}[t]
\centering
\caption{\textbf{Training against degraded verifiers.} 150 policy-gradient
steps from the same SFT initialisation, held-out-gate greedy evals, and
behavioural signatures as first-to-last 20-step means.
Proxy$^{\mathrm{train}}_{150}$ and Cited are
training-time values; the text quotes held-out proxies and cited counts at
$n{=}\NTermEval$ instead, which differ. The harmful gap types produce
distinct pathologies: padding (sufficiency), saturation starvation
(outcome-only) and dilution (noise).}
\label{tab:training}
\resizebox{\textwidth}{!}{% AUTO-GENERATED by experiments/make_macros.py -- do not edit.
\begin{tabular}{lccccccc}
\toprule
Training verifier & True$_0$ & True$_{150}$ ($n{=}96$) & Proxy$^{\mathrm{train}}_{150}$ & Gap & Starved & Cited & Words \\
\midrule
Coarse taxonomy (payload) & 0.731 & 0.844 $\pm$ 0.017 & 0.961 & 0.017 & 9/150 & 6.2$\to$4.2 & 61$\to$45 \\
Coarse taxonomy (s12) & 0.731 & 0.806 $\pm$ 0.022 & 0.975 & 0.000 & 12/150 & 4.0$\to$5.4 & 64$\to$47 \\
Exact (control, s11) & 0.731 & 0.838 $\pm$ 0.019 & 0.932 & 0.000 & 11/150 & 5.3$\to$3.9 & 65$\to$54 \\
Exact (control, s12) & 0.731 & 0.905 $\pm$ 0.016 & 0.887 & 0.000 & 24/150 & 4.0$\to$3.8 & 67$\to$50 \\
Stale policy (s11) & 0.731 & 0.875 $\pm$ 0.016 & 0.619 & -0.314 & 16/150 & 5.0$\to$3.3 & 62$\to$41 \\
Stale policy (s12) & 0.731 & 0.840 $\pm$ 0.021 & 0.458 & -0.375 & 4/150 & 4.1$\to$4.5 & 66$\to$37 \\
LLM judge & 0.731 & 0.865 $\pm$ 0.018 & 1.000 & 0.036 & 26/150 & 5.3$\to$4.2 & 70$\to$51 \\
Extraction noise $\varepsilon{=}.05$ & 0.731 & 0.821 $\pm$ 0.015 & 0.814 & -0.031 & 0/150 & 4.7$\to$3.8 & 67$\to$55 \\
Extraction noise $\varepsilon{=}.15$ & 0.731 & 0.678 $\pm$ 0.024 & 0.515 & -0.154 & 0/150 & 5.8$\to$4.3 & 71$\to$51 \\
Extraction noise $\varepsilon{=}.15$ (s12) & 0.731 & 0.648 $\pm$ 0.029 & 0.346 & -0.162 & 0/150 & 3.9$\to$4.5 & 61$\to$50 \\
Extraction noise $\varepsilon{=}.30$ & 0.731 & 0.641 $\pm$ 0.026 & 0.456 & -0.199 & 0/150 & 5.1$\to$4.5 & 66$\to$71 \\
Outcome-only & 0.731 & 0.714 $\pm$ 0.024 & 1.000 & 0.233 & 137/150 & 6.2$\to$4.6 & 73$\to$54 \\
Recall-only & 0.731 & 0.692 $\pm$ 0.017 & 1.000 & 0.470 & 75/150 & 6.1$\to$7.4 & 70$\to$60 \\
Sufficiency proxy (s11) & 0.731 & 0.667 $\pm$ 0.018 & 1.000 & 0.458 & 68/150 & 6.0$\to$7.5 & 65$\to$57 \\
Sufficiency proxy (s12) & 0.731 & 0.639 $\pm$ 0.020 & 0.988 & 0.560 & 51/150 & 5.1$\to$8.6 & 68$\to$58 \\
\bottomrule
\end{tabular}
}
\end{table}

\paragraph{Optimistic proxies are hacked by padding.}
Under $\tilde{V}_{\mathrm{suff}}$ the proxy's terminal report is
\ProxyTermSuffRun{} while true reward falls from \EvalTrueStart{} to
\TrueEndSuffRun{}, a terminal gap of \GapTermSuffRun{} on the
$n{=}\NTermEval$ held-out set. These runs and the recall-only cell below are
the only ones whose citation counts inflate beyond the initialisation band,
reaching \CitedLastSuffRun{} and \CitedLastSuffRunB{} features here, which is
the signature we predicted from correlational data before the cell ran.
Sufficiency cannot see padding. A late-training hacked explanation cites
eight features where three carry the decision, which makes the cited set
trivially sufficient (Appendix~\ref{app:qualitative} gives the rollouts
verbatim). Across the grid (Fig.~\ref{fig:signatures},
Appendix~\ref{app:transfer}) the sufficiency seeds' citation trajectories
leave every other cell within the first third of training while their word
counts stay unremarkable, so the hack targets the quantity the proxy cannot
see and leaves length alone. A pre-registered recall-only proxy, optimistic
by a different mechanism, does the same, ending at \ProxyTermRecallRun{}
against a true \TrueEndRecallRun{} with citations rising from
\CitedFirstRecallRun{} to \CitedLastRecallRun{} (Table~\ref{tab:training}).
What gets exploited is optimism about a behaviour the policy can reach. The
sufficiency formula is incidental.

\paragraph{Outcome-only rewards starve training.} Under
$\tilde{V}_{\mathrm{out}}$ the proxy saturates at \ProxyEndOutcomeRun{}
almost immediately, because every rollout carries the right decision word,
and group-relative advantages vanish. \SkipsOutcomeRun{} of 150 steps yield
zero learning signal, while true
reward drifts to \TrueEndOutcomeRun{} $\pm$ \TrueEndSEOutcomeRun. Starvation is an interaction between verifier and trainer. Any
group-relative trainer zeroes its advantage when a binary reward saturates,
the symptom dynamic sampling treats \citep{yu2025dapo}, and a binary
outcome check saturates as soon as the decision word is right, long before
the reasons are.

\paragraph{Noise dilutes, given a stable optimizer.}
Random verifier noise is reportedly benign up to $15\%$ for outcome-level
rewards \citep{imperfectverifier2026}. Per-feature noise at that rate
corrupts $1-(1-\varepsilon)^d \approx 0.80$ of rollouts ($d{=}10$), yet
under clipping the damage is a saturating dose-response
(Fig.~\ref{fig:dose}). True
reward ends at \TrueEndNoiseRunLo{} for $\varepsilon{=}.05$, close to
control level, then \TrueEndNoiseRun{} at $.15$ and \TrueEndNoiseRunHi{} at
$.30$, while the noisy verifier's own report falls faster than the truth it
corrupts. The trained policy is better than its verifier believes. Two
caveats make this a conditional safety result. The same $\varepsilon{=}.15$
cell collapsed to $0.083$ under an unclipped optimizer
(Appendix~\ref{app:grpo}), so noise budgets compose with optimizer
robustness. And noise is the one gap with zero skips, so it wastes some signal on
every step but never starves one.

\paragraph{Ranking-preserving gaps are harmless.} Stale-policy verifiers, with one rule deleted from the verifier's
model of the gate, end at \TrueEndFamdropRun{} and \TrueEndFamdropRunB{}
across seeds, both inside the control seed spread. A coarsened feature
taxonomy ends at
\TrueEndCoarseRun{} and \TrueEndCoarseRunB{} across seeds. These proxies are
wrong about levels but nearly right about rankings \citep{skalse2023invariance}
(coarse audit selection error \SelErrCoarsePayload{}), and ranking is most of
what the group-relative gradient consumes. The coarse cell is the one harmless-band
prediction that seed 12 breaks, costing \CoarseDamB{} against a band of
$0.05$, which we report rather than average away.

The LLM-judge cell (claude-haiku as extractor) went against our informal
prediction and with our formal one. Its training-time proxy
was optimistic (\EvalProxyStartJudgeRun{} against true \EvalTrueStart{} at
step 0), and from that we expected hacking. The audit said otherwise. On
the training distribution the judge has selection error \SelErrFamJudge{}
with signed gap \SignedFamJudge{}, a pessimistic level shift with little
ranking corruption, because its citation judgements closely track the
deterministic extractor. The cell ended at \TrueEndJudgeRun{}, level with the controls
(\ControlTrue). In this one cell, a judge that passed a held-out exact audit
did no measurable harm as a process verifier.

\subsection{Damage as a function of the audit}
\label{sec:law}

\begin{figure}[t]
\centering
\includegraphics[width=\textwidth,height=0.78\textheight,keepaspectratio]{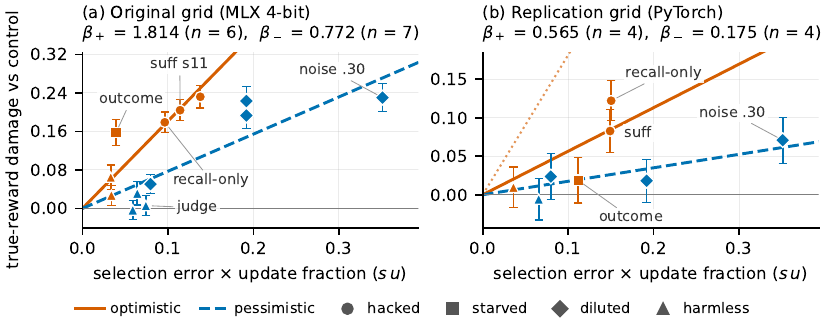}
\caption{\textbf{The gap law on two grids.} Terminal true-reward damage
against the audit's selection error weighted by the effective-update
fraction, with standard errors at $n{=}\NTermEval$. Optimistic gaps
(vermillion) lie on a steeper through-origin line than pessimistic ones
(blue). (a) is the MLX grid, all \NLawCells{} non-control cells; (b) is the
PyTorch grid fit on its own. The asymmetry reproduces, the slopes do not,
and the dotted line in (b) is the MLX optimistic slope on that panel's
scale. Only cells the text argues about are named;
Table~\ref{tab:lawpoints} lists every point.}
\label{fig:law}
\end{figure}

The phenotypes compress into one relation between audit and damage
(Fig.~\ref{fig:law}a). Let $s$ be
the verifier's selection error audited on the training distribution of
gates (the probe suite of \S\ref{sec:audit} run on 60 sampled family
episodes, Appendix~\ref{app:lawdata}, which is what lets the stale-policy
and judge verifiers be audited at all) and $u$ the fraction of training
steps with non-zero group variance. Then
\[
\text{damage} \;\approx\; \beta_{\pm}\, s\,u, \qquad
\beta_{+} = \SlopeOpt \;\; (R^2 = \RTwoOpt,\; n = \NOpt), \quad
\beta_{-} = \SlopePess \;\; (R^2 = \RTwoPess,\; n = \NPess),
\]
with a two-branch $R^2$ of \RTwoPooled{} over all \NLawCells{} cells. The
sign is the audit's signed gap, and verifiers that overestimate quality
damage $\SlopeRatio\times$ more per unit of ranking corruption reaching the
gradient than verifiers that underestimate it. The mechanism is that an
optimistic, ranking-corrupting verifier rewards a reachable
degenerate behaviour, so the policy optimises toward it, while a pessimistic
one mislabels good behaviour and wastes signal. Both regressors are
measurable before the run they predict, $s$ from a probe audit with no policy
model and $u$ from a short pilot of group variance.

This fit describes one grid. The training-distribution audit was run after
that grid had trained and before any replication run, so the fit is
retrospective here and predictive there, with the refit that scored the
replication documented in Appendix~\ref{app:second-model}. Bootstrap 95\%
intervals are
$\beta_+ \in \SlopeOptCI$ and $\beta_- \in \SlopePessCI$, and the ratio's
interval is \SlopeRatioCI{} with $P(\beta_+ > \beta_-) = \PRatioGtOne$.
Leave-one-out prediction has mean absolute error \LooMAE{}, and the law
under-predicts the starved outcome-only cell by about half. The optimistic
branch holds \NOpt{} cells from four different verifiers
(Appendix~\ref{app:lawdata} tabulates every point). The constants did
not survive replication on an independent
implementation, whereas the sign and the ordering did, and the size of the
asymmetry is at least not contradicted (\SlopeRatio$\times$ here, interval
\SlopeRatioCI, against \RepSlopeRatio$\times$ there, interval
\RepSlopeRatioCI{}, which is too wide to call agreement). We therefore state
the law as an ordering and a sign asymmetry with an interval, and claim no
portable coefficients.

\section{Replication on a Second Implementation}
\label{sec:replication}

\begin{figure}[t]
\centering
\includegraphics[width=\textwidth,height=0.78\textheight,keepaspectratio]{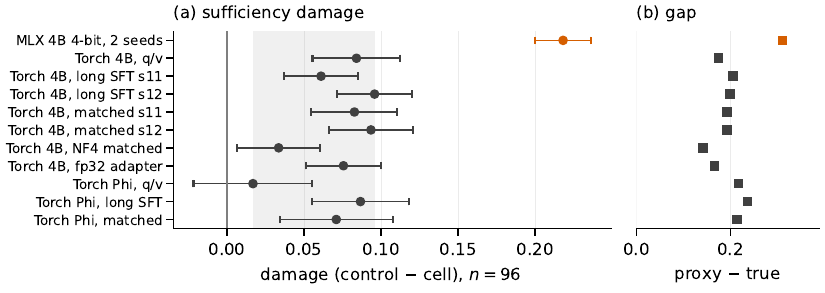}
\caption{\textbf{Replication.} (a) Sufficiency damage, control minus cell at
$n{=}\NTermEval$, in every cell measured; Torch means PyTorch fp16 unless
NF4 or fp32 is stated. The shaded band spans the smallest to largest
PyTorch damage, and every PyTorch cell (grey) falls well below the MLX grid
(vermillion). (b) The same cells' proxy over-report
barely moves, so the audit quantity is a property of the verifier while the
damage is not.}
\label{fig:replication}
\end{figure}

The grid of \S\ref{sec:anatomy} is one framework (MLX, Apple's
machine-learning framework, with 4-bit weights), one base model and one
LoRA adapter recipe (a low-rank weight update trained in place of the full
model). This section asks which of its findings survive a second
implementation. We ported the trainer to PyTorch and re-ran the grid on an
NVIDIA T4 under eight further configurations. They vary the model (Qwen3-4B and Phi-3.5-mini), the SFT length (400 and 1200
iterations), the adapter (query/value only, or every projection in the last
16 layers as in MLX), the base precision (fp16 and NormalFloat quantisation, NF4), the adapter
precision (fp16 and fp32) and the seed. Seven sets of predictions were
committed before their runs, two of them
amended after first contact with data (a rescoring against the
$n{=}\NTermEval$ refit, and a trainability gate for the corpus), and
Appendix~\ref{app:second-model} scores each as it fell.

\paragraph{What replicates.} The audit quantity is a property of the verifier
alone. The sufficiency proxy over-reports truth by between \RepGapMin{} and
\RepGapMax{} regardless of model, framework, adapter, precision or
initialisation (Fig.~\ref{fig:replication}b).

The four phenotypes replicate on the two PyTorch grids that ran every
verifier. The sufficiency cell's damage is positive in all \RepNTorchCells{} PyTorch
cells, beyond two of its own standard errors in \RepNTorchSig{}, and null on
the weakest initialisation. Every sufficiency
cell inflates its citation count, and under the matched adapter
the stale policy does \RepMFamDam{} and the coarse taxonomy
\RepMCoarseDam{}, both inside the pre-registered harmless band of $0.05$.
The remaining cells are unordered within their standard errors. The
outcome-only proxy saturates at \RepMOutProxy{} and skips \RepMOutSkips{}
of 150 matched-adapter updates. Each matched cell was re-run at a second
seed, and \SeedCellsWithin{} of \SeedCellsDone{} damages fall within two
combined standard errors. One pair is nonetheless a warning: the starved
outcome-only cell lost \SeedBOutcomeDam{} at seed 12, as much as sufficiency
(\SeedBSuffDam), with its proxy still saturated, so a starved run drifts
where the verifier cannot see.

\paragraph{The sign asymmetry replicates.} The recall-only verifier of
\S\ref{sec:anatomy} was pre-registered on this grid too, and hacks harder
than sufficiency does, at \RepRecallDam{} $\pm$ \RepRecallDamSE{} of damage
(\RepRecallSigma$\sigma$) against sufficiency's \RepTQMDam{} under the same
adapter and seed, with citation inflation \RepRecallInfl{}, the largest of
any PyTorch cell. Where both were run, the optimistic verifier without a
precision term is the worse one. Fitting the PyTorch cells alone gives $\beta_+ = \RepSlopeOpt$ ($n = \RepNOpt$, $R^2 =
\RepRTwoOpt$) and $\beta_- = \RepSlopePess$ ($n = \RepNPess$, $R^2 =
\RepRTwoPess$), a ratio of \RepSlopeRatio{} against \SlopeRatio{} on the
original grid (Fig.~\ref{fig:law}b). Its own interval is \RepSlopeRatioCI{}
with $P(\beta_+ > \beta_-) = \RepPRatioGtOne$, wide enough that only the
sign, not the size, replicates here.

\paragraph{What does not.} The size of the damage does not replicate. \RepNTorchCells{} PyTorch cells put
sufficiency damage between \RepTorchDamMin{} and \RepTorchDamMax{}, while the MLX grid
gives \RepMlxDam{} $\pm$ \RepMlxDamSE{} (Fig.~\ref{fig:replication}a). None
of the obvious explanations, each pre-registered and tested in turn,
accounts for it. Adapter capacity does not: matching the MLX adapter moved
damage from \RepTQDam{} at 1.3M parameters to \RepTQMDam{} at 7.3M. Base
quantisation does not (NF4 weights, \RepTQFourDam{} $\pm$ \RepTQFourDamSE{}),
nor does adapter precision (fp32, no loss scaling, \RepTQFpDam{} $\pm$
\RepTQFpDamSE{}), nor does more optimisation, since a stronger Qwen
initialisation gave \RepTQLDamMean{} over two seeds and 300 Phi steps changed
nothing (\PhiThreeDam{} $\pm$ \PhiThreeDamSE{}). A larger model did not fit the hardware
(Appendix~\ref{app:second-model}). One prediction in this series held:
raising Phi's initialisation from \RepTPCtrl{} to \RepTPLCtrl{} turned a
null (\RepTPDam) into a hack (\RepTPLDam, $2.8\sigma$) with nothing else
changed, so competence gates whether the hack bites. What sets how hard it
bites remains open, with three explanations excluded.

\section{Transfer to Real Policies}
\label{sec:transfer}

This section tests what carries over from synthetic gates to real
policies, first as a transfer test and then as a training domain. Models
trained only on synthetic gates must explain decisions of real,
held-out policies whose schemas they have never seen
(Appendix~\ref{app:corpus}), scored against each policy's verified logical
oracle over 80 episodes. The one significant result is damage export. The
sufficiency-hacked model's padding transfers intact, with recall
\TransSuffRec{} (it cites nearly every feature) at a precision of
\TransSuffPrec{} that ties SFT for the lowest of any variant, yet its
aggregate $F_1$ of \TransSuffFOne{} is the highest in the table, because
indiscriminate citation is $F_1$-favourable here. In this test the hacked
model's damage survived domain shift, and the aggregate score hid it.

Exact-reward training itself
transfers without aggregate loss ($F_1$ \TransExactAFOne{} and
\TransExactBFOne{} against a base of \TransBaseFOne{}) and moves precision
the right way (\TransExactAPrec{} against \TransBasePrec{}), but every
$F_1$ interval overlaps every other and the precision shift is not
significant (\PermExactPrecP), so it is directional only. Gate coverage
gives a null. A variant trained on a single gate (structural
coverage \CovSingleStruct{} against \CovFamStruct{}) transfers just as well
(\TransSingleFOne{}), so causal-precision pressure carries transfer here and
gate diversity adds nothing measurable (Appendix~\ref{app:transfer}).

\paragraph{The corpus as a second training domain.} We also trained the
four-verifier grid on the corpus itself (Appendix~\ref{app:corpus-grid}).
The audit transferred before any training, outcome-only again a coin flip
(selection error \CorpSelErrOutcome) and sufficiency again optimistic. Training
was starved: real minimal reasons are one or two features, the base model
already scores \CorpBase{} on them, and the exact control skipped
\CorpCtrlSkips{} of 150 updates, failing the pre-registered trainability gate.
The sufficiency cell is again the largest, but no corpus damage is
distinguishable from zero. This is our own starvation phenotype arriving
through task ease, and we do not count it as a replication.

\section{Recommendations}
\label{sec:discussion}

Within the scope tested here (\S\ref{sec:limits}), our results support
four recommendations. Audit a process verifier's selection error on a probe
grid before training against it. In our grids it predicted the phenotype
where reward accuracy did not, and it needs no policy model. Read the sign
too: optimistic verifiers were hacked and pessimistic ones only diluted, so
bias pessimistic when forced to choose. Do not train a process objective on
an outcome-only signal, which is blind to reasons and, once saturated,
silent. Give sufficiency- and recall-style rewards a precision term, since
alone they reward padding. An LLM judge can be used once its rankings pass a
held-out exact audit, as the one judge we trained against did.

\section{Limitations}
\label{sec:limits}
Exactness is bought by a restriction to first-match and deny-overrides
(Cedar-style, where a matching forbid overrides any permit) gates over categorical contexts. The quantitative fit belongs to one grid
(one task family, one 4B model, one recipe, MLX 4-bit), its constants did
not transfer, that grid's damage size is unexplained, and its outcome-only,
judge, recall-only and two of three noise cells are single-seed. No model
above 4B was trained, since an 8B model in 4-bit did not fit a T4. The
deterministic extractor anchors truth, audited by a held-out LLM extractor,
not humans. Corpus training was starved, so the corpus gives an audit and a
transfer test but no damage measurement, and ported policies abstract some
constructs (Appendix~\ref{app:corpus}).

\section*{Broader Impact Statement}
This work studies failure modes of reward verifiers using synthetic policy
gates and public, permissively licensed policy corpora. No human subjects,
personal data or proprietary data are involved. Understanding how proxy
rewards are hacked could in principle inform adversaries, but the attacks
demonstrated (padding, saturation exploitation) are already folklore. The
work contributes a defence, an audit that detects a dangerous verifier before
it is trained against.

\section*{Reproducibility Statement}
Every number in this paper is generated from committed result files by a
single script. No number in any table or figure is hand-typed, and the few
narrative literals restate macro values or logged training printouts. The
full pipeline (oracle, verifiers, probe audit, trainer with per-step
dual-reward logs, corpus porting protocol with per-policy provenance,
Kaggle kernel and figure generation) will be released as one self-contained
repository, including the episode data and the SFT initialisation every run
starts from, so every figure and table regenerates from a clone. The
padding-signature prediction for the sufficiency cell was written before
that cell ran but was not timestamped, so we describe it as predicted rather
than pre-registered. The seven replication pre-registrations
(Appendix~\ref{app:second-model}) are committed, timestamped files in the
artifact. The MLX grid ran on a single Apple-silicon laptop (4B model, LoRA,
about 30 hours). The replication grids ran on single NVIDIA T4 sessions.

\section*{AI Use Statement}
LLMs appear in this work as subject models (the trained 4B policies and the
frontier models in the prompting study), as one of the audited verifiers
(the LLM-judge cell) and as the held-out extraction auditor. Beyond that,
LLM assistance was limited to grammatical checks on author-written text and
to coding assistance. No claim, result, figure or citation in this paper was
generated by an LLM, and all experimental results derive from the released
deterministic pipeline.

\bibliography{references}
\bibliographystyle{tmlr}

\appendix

\section{The Exact Oracle}
\label{app:oracle}

\paragraph{Tensor oracle.} For enumerable $\mathcal{X}$ the gate
materialises as an int8 decision tensor $T$ of shape
$(12,3,4,2,2,2,3,3,4,3)$. A set $S$ is sufficient at $c$ iff the tensor
slice obtained by pinning $c$'s values on $S$ is constant, which is one
\texttt{min==max} reduction. MSR enumeration walks the $2^{10}$ subsets in
popcount order with superset pruning, so the survivors are exactly the
minimal sufficient sets in the sense of \citet{darwiche2020reasons}, and
decisive features use single-axis slices. Formal XAI of this kind buys
soundness guarantees that heuristic attribution does not
\citep{marquessilva2022delivering}.
Exactness is by construction, because every counterfactual completion is
literally enumerated.

\paragraph{Logical oracle.} For non-enumerable or foreign-schema gates the
same queries are decided by abduction over rule-relevant features only. For
first-match block-rule gates, BLOCK-sufficiency is an adversarial CSP (can
the unpinned features be completed so that no rule matches?) solved by
backtracking, the same reduction production cloud authorization uses to decide
policy questions by SMT \citep{backes2018zelkova}, and ALLOW-sufficiency requires every rule to be falsified by a
pinned literal. Deny-overrides (Cedar-style) gates add permit and forbid
reachability checks. On the enumerable schema, the logical oracle's decisions
and full MSR sets agree with the enumerating tensor oracle on
\OracleChecked{} (script and result file in the artifact).

\section{Degraded Verifiers: Definitions and Invariants}
\label{app:dial}

Let $E(t)$ be the deterministic extractor's cited set and $r^\star$ the
$F_1$-based exact reward of Definition~2. Each degraded verifier $\tilde V$
replaces exactly one component. \textbf{Outcome-only} scores
$\mathbf{1}[\text{decision word correct}]$ minus the shared penalties.
\textbf{Sufficiency} scores the graded sufficiency fraction of $E(t)$ by
pin-and-check on the episode's own gate tensor. \textbf{Recall-only} scores
$\max_{m \in \mathrm{MSR}(c)} |E(t) \cap m|/|m|$ minus the shared penalties,
with no precision term. \textbf{Noise}$(\varepsilon)$ applies i.i.d.\
per-feature flips to $E(t)$, seeded by $(\text{seed},c,t)$ so that rewards
are reproducible. \textbf{Coarse}$(G)$ scores precision and recall at the
granularity of feature groups $G$. \textbf{Stale}$(r)$ recomputes decision
and MSRs on the gate with rule $r$ deleted. \textbf{Judge} replaces $E$ with
an LLM extractor, with the deterministic $E$ held out as the audit; optimising
against a model that reads explanations is the setting in which
\citet{baker2025monitoring} observe obfuscation under strong pressure. Each
satisfies a designed-rate invariant enforced by unit test. Noise$(0)$ is
identical to exact and noise$(1)$ flips every feature, coarse with singleton
groups is identical to exact, and stale$(r)$ equals exact on every context
whose decision and MSRs rule $r$ does not touch. Table~\ref{tab:audit-full}
reports every verifier's measured selection error and signed gap on the
probe grid, including all eight stale-policy variants.

\begin{table}[H]
\centering
\caption{The full model-free gap audit, one row per stale-policy rule.}
\label{tab:audit-full}
% AUTO-GENERATED by experiments/make_macros.py -- do not edit.
\begin{tabular}{llccc}
\toprule
Verifier & Gap type & Reward gap & Selection error & Decision disagr. \\
\midrule
Exact (control) & n/a & 0.000 & 0.000 & n/a \\
Outcome-only & metric & 0.629 & 0.504 & n/a \\
Sufficiency proxy & metric & 0.239 & 0.208 & n/a \\
Recall-only proxy & metric & 0.090 & 0.132 & n/a \\
Extraction noise $\varepsilon{=}.05$ & extraction & 0.051 & 0.043 & n/a \\
Extraction noise $\varepsilon{=}.15$ & extraction & 0.156 & 0.143 & n/a \\
Extraction noise $\varepsilon{=}.30$ & extraction & 0.271 & 0.308 & n/a \\
Coarse taxonomy (payload) & granularity & 0.034 & 0.030 & n/a \\
Coarse taxonomy (session) & granularity & 0.076 & 0.048 & n/a \\
Stale policy (drop R1) & policy model & 0.035 & 0.026 & 0.0252 \\
Stale policy (drop R2) & policy model & 0.026 & 0.023 & 0.0261 \\
Stale policy (drop R3) & policy model & 0.040 & 0.024 & 0.0237 \\
Stale policy (drop R4) & policy model & 0.057 & 0.042 & 0.1011 \\
Stale policy (drop R5) & policy model & 0.060 & 0.032 & 0.0476 \\
Stale policy (drop R6) & policy model & 0.054 & 0.040 & 0.0425 \\
Stale policy (drop R7) & policy model & 0.031 & 0.023 & 0.0229 \\
Stale policy (drop R8) & policy model & 0.037 & 0.029 & 0.0611 \\
\bottomrule
\end{tabular}

\end{table}

\section{Real-Policy Corpus: Porting Protocol and Provenance}
\label{app:corpus}

\begin{table}[H]
\centering
\caption{\textbf{Real-policy corpus v2.} \NCorpusPolicies{} public policies
ported to executable gates under a fidelity-first protocol, with every
abstraction documented and sanity contexts hand-traced against the original
semantics. \NCorpusHeldout{} policies are held out entirely.}
\label{tab:corpus}
\resizebox{\textwidth}{!}{% AUTO-GENERATED by experiments/make_macros.py -- do not edit.
\begin{tabular}{llccccc}
\toprule
Policy & Source & Features & $|\mathcal{X}|$ & Rules & Block rate & Split \\
\midrule
\texttt{cedar-photoapp} & cedar-examples & 8 & 384 & 5 & 0.43 & train \\
\texttt{cedar-streaming-service} & cedar-examples & 10 & 3{,}456 & 7 & 0.73 & heldout \\
\texttt{cedar-tags-n-roles} & cedar-examples & 5 & 162 & 2 & 0.80 & train \\
\texttt{cedar-tax-preparer} & cedar-examples & 7 & 128 & 4 & 0.81 & heldout \\
\texttt{gk-block-endpoint-edit} & gatekeeper-library & 5 & 112 & 2 & 0.11 & train \\
\texttt{gk-container-limits} & gatekeeper-library & 5 & 192 & 8 & 0.48 & heldout \\
\texttt{gk-disallowed-tags} & gatekeeper-library & 4 & 72 & 3 & 0.25 & train \\
\texttt{gk-psp-hostnet} & gatekeeper-library & 5 & 64 & 3 & 0.22 & train \\
\texttt{gk-psp-seccomp} & gatekeeper-library & 3 & 36 & 3 & 0.25 & heldout \\
\texttt{gk-psp-users} & gatekeeper-library & 6 & 432 & 7 & 0.15 & train \\
\texttt{gk-required-labels} & gatekeeper-library & 4 & 60 & 4 & 0.70 & train \\
\texttt{gk-storageclass} & gatekeeper-library & 5 & 64 & 7 & 0.75 & train \\
\texttt{opa-loadbalancer} & library & 3 & 40 & 4 & 0.78 & heldout \\
\texttt{opa-pvc-resize} & library & 3 & 54 & 1 & 0.81 & train \\
\bottomrule
\end{tabular}
}
\end{table}

Each public policy (Table~\ref{tab:corpus}) is ported by mapping every Cedar
\citep{cutler2024cedar} permit or forbid clause, or Rego violation rule
\citep{opa,gatekeeper}, to one antecedent over categorical features. Entity-relationship conditions become boolean
features, and numeric, regex and datetime conditions are pre-bucketed into
feature values. Decision semantics are preserved exactly under
\textsc{DenyOverrides} (default deny, permits allow, forbids override) or
admission style (empty permit plus violation forbids). Every abstraction is
recorded in a per-policy \texttt{omitted} field, and every port was
validated by exhaustive enumeration against the gate plus hand-traced sanity
contexts against the original source semantics, three or more per policy,
all matching. Full machine-readable specs with provenance (repository, path,
license, block rate) ship in the artifact, and all sources are Apache-2.0.
The per-policy abstraction record below is generated verbatim from those
\texttt{omitted} fields.

{\small
% AUTO-GENERATED by experiments/make_macros.py -- do not edit.
\sloppy
\begin{description}\setlength{\itemsep}{2pt}
\item[\texttt{cedar-photoapp}] (\texttt{cedar-policy/\allowbreak{}cedar-examples}: \texttt{cedar-policy-language-in-action/\allowbreak{}PhotoApp/\allowbreak{}photoapp.cedar}, Apache-2.0). The scope condition \textasciigrave{}resource in PhotoApp::Application::''PhotoApp''\textasciigrave{} in every policy treated as always-true (all Photos are in the Application per the entity store). \textasciigrave{}resource has subjects \&\& resource.subjects.contains(principal)\textasciigrave{} in @id('Photo.subjects') bucketed into is\_subject. \textasciigrave{}context has judgingSession \&\& context.judgingSession == true\textasciigrave{} in @id('PhotoJudge') bucketed into judging\_session (absent attribute to no). The instance policies @id('DoeFamily') (DoeFamily members on Album DoePhotos) and @id('JaneVacation') (JohnDoe on Album JaneVacation) merged into one relationship boolean album\_grant (permits are disjunctive, so merging preserves decisions). @id('Photo.owner')'s explicit action list is all three schema actions, so the conjunct is dropped as vacuous. Labels is a Set\textless{}String\textgreater{}, so label\_private/\allowbreak{}label\_contest are independent booleans.
\item[\texttt{cedar-streaming-service}] (\texttt{cedar-policy/\allowbreak{}cedar-examples}: \texttt{cedar-example-use-cases/\allowbreak{}streaming\_service/\allowbreak{}policies.cedar}, Apache-2.0). All datetime arithmetic bucketed: now-vs-resource.releaseDate in @id('subscriber-content-access/\allowbreak{}show') and @id('early-access-show') to time\_window (meaningful only for shows; Movie has no releaseDate, so the value is a don't-care never consulted for movies). The conjunction resource.isOscarNominated \&\& 2025-02-02T19:00-0500 \textless{}= now \textless{} 2025-03-02T19:00-0500 in @id('rent-buy-oscar-movie') to single boolean oscar\_promo (a 4th 'oscar\_window' time\_window value was rejected: shows at Oscars-window times still have a definite release relation, so that enum would create ill-defined contexts and break exactness). The toTime()/\allowbreak{}localTimeOffset computation in the unless-clause of @id('forbid-bedtime-watch-kid-profile') to is\_bedtime (yes iff local time outside 06:00-21:00). FreeMember/\allowbreak{}Subscriber entity types and subscription.tier merged into 3-valued membership (FreeMember has no tier or profile; no rule consults profile\_is\_kid for free members). The @id('subscriber-content-access/\allowbreak{}show') unless-clause (A \&\& B) split into two permits by De Morgan (not-early OR after-release). Needs\_rent\_or\_buy /\allowbreak{} is\_oscar\_nominated are Movie-only and is\_early\_access is Show-only attributes kept as global features (don't-cares for the other type).
\item[\texttt{cedar-tags-n-roles}] (\texttt{cedar-policy/\allowbreak{}cedar-examples}: \texttt{cedar-example-use-cases/\allowbreak{}tags\_n\_roles/\allowbreak{}policies.cedar}, Apache-2.0). Principals holding multiple roles simultaneously (entity Joe is in both Role-A and Role-B with different tag sets; Cedar's decision is the disjunction over per-role policies), each context models the evaluation for a single role membership, which is exact for single-role principals. The guard \textasciigrave{}principal.allowedTagsForRole has ''Role-X''\textasciigrave{} repeated in every when-clause of @id('Role-A policy')/\allowbreak{}@id('Role-B policy') is folded into the match enums (a missing role record maps every dimension to not\_covered, which is a consistent assignment). Per-dimension set logic bucketed into 3-valued enums: principal\_all = the user's role record lacks the tag OR its values contain 'ALL' (user-side wildcard, matches any resource). Covered = resource lacks the tag, resource values contain 'ALL', or user values are a superset of resource values (containsAll). Not\_covered = the when-clause for that dimension is false. The Role-A action conjunct is kept explicit even though Role-A Actions contains all three schema actions.
\item[\texttt{cedar-tax-preparer}] (\texttt{cedar-policy/\allowbreak{}cedar-examples}: \texttt{cedar-example-use-cases/\allowbreak{}tax\_preparer/\allowbreak{}policies.cedar}, Apache-2.0). The only action is viewDocument, so no action feature. Rule 1a's assigned\_orgs.contains(\{organization: resource.owner.organization, serviceline: resource.serviceline, location: resource.location\}) set-of-records containment bucketed into assignment\_covers\_doc. The organization/\allowbreak{}serviceline/\allowbreak{}doc\_location document descriptors are kept for realism but are decision-irrelevant (only the tuple-match boolean is consulted, since a professional may hold several assigned\_orgs tuples). The @id('adhoc-access') template and its \textasciigrave{}linked\textasciigrave{} instantiation (link\_id AliceView: Alice to Document DEF) generalized to the boolean adhoc\_grant permit. Rule 2's context.consent record split into consent\_from\_owner (consent.client == resource.owner) and consent\_covers\_location (team\_region\_list.contains(principal.location)). Rule 2's \textasciigrave{}unless (A \&\& B)\textasciigrave{} becomes two forbids (not-A; not-B).
\item[\texttt{gk-block-endpoint-edit}] (\texttt{open-policy-agent/\allowbreak{}gatekeeper-library}: \texttt{library/\allowbreak{}general/\allowbreak{}block-endpoint-edit-default-role/\allowbreak{}template.yaml}, Apache-2.0). Context is a single (role, policy-rule, verb) grounding of the rego's iteration over rules[\_], resources[\_], verbs[\_]. The three hasEditVerb helper bodies collapse into the value set \{create,patch,update\}. Exact string match on metadata.name 'system:aggregate-to-edit' bucketed into role\_name. Object\_kind and has\_autoupdate\_annotation are faithful distractors, the rego never inspects kind (ClusterRole scoping is in the Constraint match) and the rbac.authorization.kubernetes.io/\allowbreak{}autoupdate annotation appears only in the advisory violation message, never in a condition.
\item[\texttt{gk-container-limits}] (\texttt{open-policy-agent/\allowbreak{}gatekeeper-library}: \texttt{library/\allowbreak{}general/\allowbreak{}containerlimits/\allowbreak{}template.yaml}, Apache-2.0). Parameters instantiated to cpu='2', memory='1Gi' (both finite, so the cpu != '-1' disable-guards are always satisfied). Canonify\_cpu/\allowbreak{}canonify\_mem numeric parsing and the \textgreater{} max comparisons are pre-bucketed into cpu\_limit/\allowbreak{}mem\_limit values (under\_max/\allowbreak{}over\_max/\allowbreak{}unparseable, unset = key absent or ''). The two rego rules 'not container.resources' and 'not container.resources.limits' merge into the has\_limits=none bucket. The two field-wrapper violation rules (containers/\allowbreak{}initContainers) are represented by leaving container\_field unconstrained in every forbid. ExemptImages prefix matching bucketed into exempt\_image. Coherence: has\_limits=none implies cpu/\allowbreak{}mem unset, partial implies exactly one unset, present implies neither unset, decisions on incoherent product contexts have no rego ground truth. Ephemeral containers not checked by the template (field cannot be set).
\item[\texttt{gk-disallowed-tags}] (\texttt{open-policy-agent/\allowbreak{}gatekeeper-library}: \texttt{library/\allowbreak{}general/\allowbreak{}disallowedtags/\allowbreak{}template.yaml}, Apache-2.0). Parameters instantiated to tags=['latest']. Suffix matching (strings.any\_suffix\_match against ':latest') and the split-by-'/\allowbreak{}' untagged check are pre-bucketed into image\_tag: none = no colon in last segment, latest = ends with ':latest', pinned = any other tag, digest = @sha256:... references which trip neither rule (colon present, suffix differs) and are admitted. The three input\_containers union rules are represented by leaving container\_field unconstrained in both forbids. ExemptImages prefix matching bucketed into exempt\_image. Image\_registry is a faithful distractor, split(image, '/\allowbreak{}') means even a registry port colon (registry:5000/\allowbreak{}app) never masks an untagged image.
\item[\texttt{gk-psp-hostnet}] (\texttt{open-policy-agent/\allowbreak{}gatekeeper-library}: \texttt{library/\allowbreak{}pod-security-policy/\allowbreak{}host-network-ports/\allowbreak{}template.yaml}, Apache-2.0). Numeric min/\allowbreak{}max range bucketed into host\_port. Host\_port=none covers no ports and ports without hostPort. Single container with a single port modeled (source iterates all containers x ports). ExemptImages prefix matching bucketed to exempt\_image (applies only to the port check, matching the rego). UPDATE operations bypass all checks (hostNetwork is immutable). CEL variant's ports.all() differs from the rego's any-port semantics, the Rego engine is ported.
\item[\texttt{gk-psp-seccomp}] (\texttt{open-policy-agent/\allowbreak{}gatekeeper-library}: \texttt{library/\allowbreak{}pod-security-policy/\allowbreak{}seccomp/\allowbreak{}template.yaml}, Apache-2.0). Allowlist instantiations: wildcard=['*'] (short-circuits all checks including exemption), runtime\_default\_only=['runtime/\allowbreak{}default'], runtime\_and\_localhost=['runtime/\allowbreak{}default','localhost/\allowbreak{}allowed.json'] (exact file match, no localhost/\allowbreak{}* wildcard). Resolved\_profile collapses the four resolution locations (pod/\allowbreak{}container annotation, pod/\allowbreak{}container securityContext), location affects only the message. Docker\_default models the annotation-scheme string 'docker/\allowbreak{}default' (securityContext RuntimeDefault canonicalizes to an allowed variant and cannot produce it). SecurityContext-scheme parameter translation (RuntimeDefault/\allowbreak{}Localhost/\allowbreak{}allowedLocalhostFiles) folded into the same canonical allowlists. Single container modeled. No update exclusion exists in this template. CEL engine variant is decision-equivalent.
\item[\texttt{gk-psp-users}] (\texttt{open-policy-agent/\allowbreak{}gatekeeper-library}: \texttt{library/\allowbreak{}pod-security-policy/\allowbreak{}users/\allowbreak{}template.yaml}, Apache-2.0). RunAsGroup and supplementalGroups families dropped to stay \textless{}=6 features (same accept\_value logic as fsGroup; supplementalGroups additionally iterates array values). Provided\_user=root means uid 0 and is assumed outside the MustRunAs ranges. The runAsNonRoot=true escape hatch (runAsUser unset but runAsNonRoot set gives no NonRoot violation) is not modeled, unset means neither field is set. Container/\allowbreak{}pod-level securityContext precedence collapsed into the single provided value. Single container modeled (containers/\allowbreak{}initContainers/\allowbreak{}ephemeralContainers union). ExemptImages prefix matching bucketed to exempt\_image.
\item[\texttt{gk-required-labels}] (\texttt{open-policy-agent/\allowbreak{}gatekeeper-library}: \texttt{library/\allowbreak{}general/\allowbreak{}requiredlabels/\allowbreak{}template.yaml}, Apache-2.0). Parameters instantiated to labels=[\{key: owner\}, \{key: env, allowedRegex: \textasciicircum{}(dev\textbar{}staging\textbar{}prod)\$\}]. The single missing-labels violation (count(required, provided) \textgreater{} 0) is grounded to one forbid per configured label. The regex-mismatch violation cannot fire for owner (no allowedRegex, so expected.allowedRegex != '' is undefined) and its regex match on env is pre-bucketed into env\_value (dev/\allowbreak{}staging/\allowbreak{}prod match, other = present-but-mismatched). Parameters.message (message override) dropped. Extra\_labels and object\_kind are faithful distractors, the rego ignores labels beyond the required set and never inspects kind (scoping is done by the Constraint match criteria).
\item[\texttt{gk-storageclass}] (\texttt{open-policy-agent/\allowbreak{}gatekeeper-library}: \texttt{library/\allowbreak{}general/\allowbreak{}storageclass/\allowbreak{}template.yaml}, Apache-2.0). Allowlist instantiated as any = allowedStorageClasses unset/\allowbreak{}['*'] and restricted\_to\_a = ['class\_a']. Class\_exists = whether the named class is in data.inventory (don't-care when storageclass\_name is unset). 'not storageclass\_allowed(name)' negates a 2-body helper, so each badname violation family grounds to two conjunctive forbids (not-in-inventory; not-in-allowlist), overlapping where both hold. The standalone StorageClasses-not-synced violation (inventory key entirely absent) is a cluster-misconfiguration guard outside the per-object feature space and is omitted (inventory assumed synced). The badname rules also fire on unset names via negation-of-undefined, decision-equivalent to the noname forbids, folded there. Context is one volumeClaimTemplate grounding for statefulsets. Include\_classes\_in\_message is a faithful distractor (message-only parameter). ApiVersion/\allowbreak{}kind checks bucketed into object\_kind.
\item[\texttt{opa-loadbalancer}] (\texttt{open-policy-agent/\allowbreak{}library}: \texttt{kubernetes/\allowbreak{}admission/\allowbreak{}loadbalancer.rego}, Apache-2.0). Source is a library defining is\_external\_lb. Ported as deny-if-external. contains(spec.type, 'LoadBalancer') substring match bucketed to service\_type=loadbalancer. Internal\_annotation values mean the exact key=value pair from internal\_annotations is present (a key with a mismatched value buckets to none). Count(externalIPs)\textgreater{}0 bucketed to has\_external\_ips (source reads object.externalIPs, not spec.externalIPs).
\item[\texttt{opa-pvc-resize}] (\texttt{open-policy-agent/\allowbreak{}library}: \texttt{kubernetes/\allowbreak{}admission/\allowbreak{}persistentvolumeclaimresize.rego}, Apache-2.0). Allow-only source (single 'verify' conjunction), so ported as a permit-conjunction over default deny rather than admission+forbids. Plugin set is the source's literal \{GlusterFS, Cinder, AWS, GCE, RDS\} ('RDS' as written upstream, likely a typo for RBD). Numeric size comparison bucketed to size\_change. Phase string 'ClaimBound' bucketed to bound, all other phases to pending/\allowbreak{}lost.
\end{description}

}

\section{Training Details}
\label{app:training}

The policy is Qwen3-4B (4-bit, MLX) with LoRA adapters (7.34M trainable parameters),
initialised from a teacher-distilled SFT adapter. Training uses policy gradient with
group-relative advantage, group 8, temperature 0.8, at most 220 new tokens,
Adam at learning rate $5\times10^{-6}$, global-norm gradient clipping at 1.0,
per-completion gradient accumulation, 150 steps, and greedy held-out evals
every 25 steps on 8 episodes during training. Steps with zero group variance
are logged and skipped with no update. Seeds are 11 and 12.
Table~\ref{tab:hparams} lists the full configuration of both the final
recipe and the initial recipe whose fragility is analysed in
Appendix~\ref{app:grpo}. Both are archived as \texttt{config.json} alongside
every run's per-step logs and terminal adapters. One run takes about 75
minutes of wall-clock (the LLM-judge cell is slower, because its extraction
API calls dominate), and the full grid takes about 30 hours.

\begin{table}[H]
\centering
\caption{Training hyperparameters, generated from the archived run
configurations. ``Same'' means the value is shared between recipes. The
initial recipe's instability is documented in Appendix~\ref{app:grpo}.}
\label{tab:hparams}
% AUTO-GENERATED by experiments/make_macros.py -- do not edit.
\begin{tabular}{lll}
\toprule
Hyperparameter & Final recipe & Initial recipe \\
\midrule
Base model & \texttt{Qwen3-4B-4bit} (MLX) & same \\
LoRA rank / scale / dropout & 8 / 20 / 0.0 & same \\
Layers adapted & 16 & same \\
Optimizer & Adam & Adam \\
Learning rate & $5\times10^{-6}$ & $1\times10^{-5}$ \\
Gradient clipping (global norm) & 1.0 & none \\
Group size (completions/step) & 8 & 4 \\
Gradient accumulation & per completion & whole group \\
Steps & 150 & 150 \\
Sampling temperature & 0.8 & 0.8 \\
Max new tokens & 220 & 220 \\
Held-out eval & every 25 steps, $n{=}8$ & same \\
Hardware & Apple M-series, 24\,GB unified memory & same \\
Wall-clock per run & ${\approx}75$\,min & ${\approx}75$\,min \\
\bottomrule
\end{tabular}

\end{table}

\paragraph{PyTorch port.} The replication trainer is a line-by-line port of
the MLX loop to PyTorch with PEFT. It keeps the group size, learning rate, clipping, temperature, token budget,
seed handling and skip rule. Three guards are new: a first-step proof of life
(a non-zero gradient norm and a changed weight), a rollout-diversity
pre-flight that aborts a run whose policy emits one text per group, and a
circuit breaker on 25 consecutive identical groups. Adapters carry a base-model marker that the loader checks, after one
run silently loaded a Phi adapter onto Qwen through a name collision. The
matched adapter targets every attention and MLP projection in the last 16
layers with rank 8 and scale 20, which is the MLX adapter exactly. A run on
a T4 takes about 3.5 hours for 150 steps.

\paragraph{Compute budget.} Table~\ref{tab:compute} itemises all training
and evaluation compute behind the MLX grid. All on-device compute ran on one
Apple M-series laptop (24\,GB unified memory). The only
external compute for that grid is \ComputeHaikuTotal{} \texttt{claude-haiku}
API calls (\ComputeJudgeCalls{} as the judge cell's training-time reward and
\ComputeExtractCalls{} for held-out transfer extraction), all
content-address cached and shipped with the artifact. At about
\ComputeTotalHours{}\,h of total wall-clock, the MLX programme sits roughly
two orders of magnitude below typical RLVR training budgets. The replication
grids add about 110 T4-hours across the four free Kaggle accounts that ran
them.

\begin{table}[H]
\centering
\caption{Compute budget by experiment family, generated from archived run
logs and API caches. Wall-clock is measured from each run's log-file
timestamps. \ComputeNFallback{} runs whose timestamp windows were polluted
by machine sleep use the recipe-typical estimate of about 71\,min instead.
The judge cell's wall-clock is elapsed time dominated by its per-rollout
reward API calls. Transfer-eval generation took about
\ComputeGenMinutes\,min on-device, the rest of that hour being
extraction-API latency.}
\label{tab:compute}
\resizebox{\textwidth}{!}{% AUTO-GENERATED by experiments/make_macros.py -- do not edit.
\begin{tabular}{llcc}
\toprule
Component & Volume & Wall-clock & \texttt{haiku} calls \\
\midrule
Definitive grid (final recipe, excl.\ judge cell) & 14 runs $\times$ 150 steps & 17.4\,h & n/a \\
LLM-judge cell (final recipe) & 1 run $\times$ 150 steps & 16.2\,h (API-bound) & 1{,}496 \\
Coverage-null control (single gate) & 1 run $\times$ 150 steps & ${\approx}$1.2\,h & n/a \\
Pilot grid (recipe selection) & 6 runs $\times$ 50 steps & 2.9\,h & n/a \\
Initial-recipe runs (Appendix~\ref{app:grpo}) & 8 runs $\times$ 150 steps & 6.9\,h & n/a \\
Transfer evaluation (\S\ref{sec:transfer}) & 6 variants $\times$ 80 episodes & ${<}1$\,h & 7{,}580 \\
\midrule
Total & n/a & ${\approx}45$\,h & 9{,}076 \\
\bottomrule
\end{tabular}
}
\end{table}

\section{Optimizer Fragility}
\label{app:grpo}

\begin{figure}[H]
\centering
\includegraphics[width=\textwidth,height=0.78\textheight,keepaspectratio]{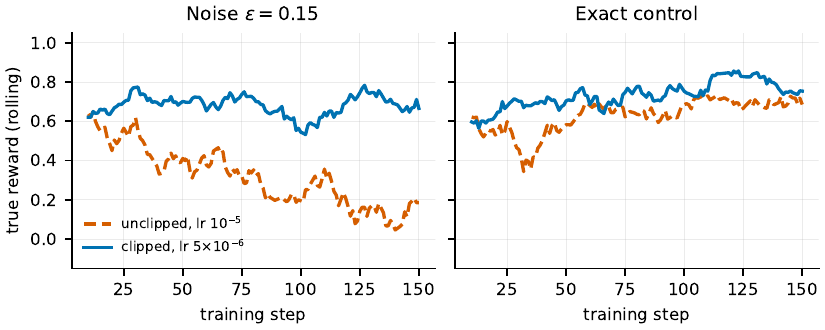}
\caption{\textbf{Optimizer fragility.} Rolling (window 10) per-step mean
true reward for the same two grid cells under the initial recipe (unclipped,
learning rate $10^{-5}$, group 4, vermillion dashed) and the final recipe
(clipped at global norm 1.0, learning rate $5{\times}10^{-6}$, group 8, blue
solid). Left, the noise $\varepsilon{=}.15$ cell collapses under the fragile
optimizer and merely dilutes under the clipped one. Right, the exact control
on seed 11 survives both (the control collapse was on seed 12), but the
unclipped trajectory dips sharply mid-run and ends below the clipped one.}
\label{fig:fragility}
\end{figure}

Two negative results shaped the recipe. First, an off-the-shelf MLX GRPO
implementation produced checkpoints byte-identical to initialisation, a
silent zero-gradient no-op. The released trainer is a transparent REINFORCE
loop with group-relative advantage, the GRPO core at on-policy single-step,
with a first-step proof-of-life assertion on gradient norms and weight
deltas. Second, the initial recipe (learning rate $10^{-5}$, group 4, no
clipping) was seed-unstable. The exact-reward control itself collapsed on
one seed to a terminal true reward of $0.401$, and under it the
noise$(\varepsilon{=}.15)$ cell collapsed to $0.083$, against
\TrueEndNoiseRun{} under the final recipe. The two old-recipe figures in
this paragraph are in-training 8-episode evaluations. Those runs predate the
terminal re-scoring and saved no adapter to re-score, so they are not
comparable at the third decimal with the $n = \NTermEval$ numbers used
everywhere else. The comparison they support is a collapse against a stable
level, which no measurement noise explains. Figure~\ref{fig:fragility} shows
both comparisons step by step: the noise cell decays roughly monotonically to
collapse under the fragile optimizer and holds a diluted level under the
clipped one, and the exact control survives both but dips sharply mid-run
without clipping. Verifier noise and optimizer fragility therefore compound:
a gap that is benign under a stable optimizer can be catastrophic under a
fragile one. All old-recipe runs are preserved in the artifact.

\section{Fit Data}
\label{app:lawdata}

The optimistic branch of the original grid holds \NOpt{} cells across four
verifiers: the sufficiency proxy at two seeds, the recall-only proxy, the
outcome-only proxy, and the coarse taxonomy at two seeds. The shape of this
relation, true reward falling as optimisation pressure is applied through a
corrupted proxy, is what \citet{karwowski2024goodhart} derive for MDPs and what
\citet{laidlaw2025correlated} formalise as a breakdown of proxy-truth
correlation over a reference policy's states; the exact oracle lets us measure
the curve rather than bound it. An earlier version
of this work reported it as one verifier's slope, before the recall-only and
outcome-only cells were added.
$R^2$ in the main text is the through-origin form
($1 - \mathrm{SS}_{\mathrm{res}}/\sum y^2$), and the standard form gives
\RTwoOptStd{} and \RTwoPessStd. Dropping $u$ and fitting selection error
alone gives $R^2 = \SOnlyOptRTwo$ optimistic and \SOnlyPessRTwo{}
pessimistic, so on this grid the model-free audit quantity carries the
prediction and the pilot-measured update fraction adds little.

Table~\ref{tab:lawpoints} lists every point behind Fig.~\ref{fig:law}a.
Per cell it gives the pre-training audit quantities (selection error $s$ and
signed reward gap, both from the model-free probe audit of
\S\ref{sec:audit}), the effective-update fraction $u$ (the share of the 150
steps with non-zero group variance, measurable from a cheap pilot), the
regressor $s\,u$, and the realised damage (control-mean terminal true reward
\ControlTrue{} minus the cell's terminal true reward). The selection errors
used in the fit come from a second audit run on the training distribution
(60 stratified episodes drawn from the sampled family gates, with the same
probe suite as Table~\ref{tab:audit}). This is what lets the stale-policy
verifier (a rule dropped from each sampled gate, which the fixed-gate audit
of Table~\ref{tab:audit} cannot represent) and the LLM judge
(\SelErrFamJudge{}, one API call per probe) enter the fit. The $s$ column
of Table~\ref{tab:lawpoints} separates the optimistic verifiers
(sufficiency \SelErrFamSuff, recall-only \RepRecallSel) from the near-chance
outcome-only verifier (\SelErrFamOutcome) and from the barely distorting
pessimistic ones (coarse \SelErrFamCoarse, stale policy \SelErrFamFamdrop,
judge \SelErrFamJudge).
The optimistic through-origin fit uses the cells with positive signed gap
and the pessimistic fit those with negative signed gap. The stale-policy
and judge cells were harmless in training, and the judge's damage is in fact
negative, with terminal true reward \TrueEndJudgeRun{} against the control
mean \ControlTrue.

Every cell is in the fit. The outcome-only run originally finished without
a terminal checkpoint, because the trainer of the day wrote adapters only on
updating steps and that cell skipped \SkipsOutcomeRun{} of its 150 updates.
It was retrained after the bug was fixed, reproducing the same
\SkipsOutcomeRun{} skipped steps and the same endpoint, and its terminal
adapter is scored here like every other. Starvation is therefore measured at
$n = \NTermEval$ on this grid as well as on the replication, where the
outcome-only proxy saturates at \RepMOutProxy{} and the matched cell loses
\RepMOutDam{} against its control.

\begin{table}[H]
\centering
\caption{Per-cell gap-law fit data behind Fig.~\ref{fig:law}a. Damage is
the control-mean terminal true reward minus the cell's, and negative damage
means the cell ended above the control mean.}
\label{tab:lawpoints}
\resizebox{\textwidth}{!}{% AUTO-GENERATED by experiments/make_macros.py -- do not edit.
\begin{tabular}{llcccccc}
\toprule
Cell & Phenotype & Sel.\ err.\ $s$ & Signed gap & Eff.\ upd.\ $u$ & $s\,u$ & Damage & True$_{150}$ \\
\midrule
Sufficiency proxy (s11) & hacked & 0.209 & +0.185 & 0.547 & 0.114 & +0.204 & 0.667 \\
Sufficiency proxy (s12) & hacked & 0.209 & +0.185 & 0.660 & 0.138 & +0.232 & 0.639 \\
Outcome-only & starved & 0.453 & +0.163 & 0.087 & 0.039 & +0.158 & 0.714 \\
Noise $\varepsilon{=}.05$ & diluted & 0.080 & -0.027 & 1.000 & 0.080 & +0.051 & 0.821 \\
Noise $\varepsilon{=}.15$ & diluted & 0.192 & -0.060 & 1.000 & 0.192 & +0.193 & 0.678 \\
Noise $\varepsilon{=}.30$ & diluted & 0.351 & -0.117 & 1.000 & 0.351 & +0.230 & 0.641 \\
Stale policy (s11) & harmless & 0.066 & -0.083 & 0.893 & 0.059 & -0.003 & 0.875 \\
Stale policy (s12) & harmless & 0.066 & -0.083 & 0.973 & 0.064 & +0.031 & 0.840 \\
Coarse taxonomy (payload) & harmless & 0.036 & +0.020 & 0.940 & 0.034 & +0.027 & 0.844 \\
LLM judge & harmless & 0.090 & -0.067 & 0.827 & 0.074 & +0.006 & 0.865 \\
Recall-only & hacked & 0.193 & +0.052 & 0.500 & 0.096 & +0.179 & 0.692 \\
Noise $\varepsilon{=}.15$ (s12) & diluted & 0.192 & -0.060 & 1.000 & 0.192 & +0.223 & 0.648 \\
Coarse taxonomy (s12) & harmless & 0.036 & +0.020 & 0.920 & 0.033 & +0.065 & 0.806 \\
\bottomrule
\end{tabular}
}
\end{table}

\section{Prompts and Probe Construction}
\label{app:prompts}

Generation uses the audit-log prompt (context rendering, full policy text,
decision, ``write 2 to 4 sentences''). Extraction uses the schema-generic
constrained-JSON prompt with the episode's feature list. Probe texts are
rendered from templates whose phrases are written against the deterministic
extractor's alias patterns, so each probe's cited set is exact by
construction. The six probe types are faithful (canonical MSR), partial
(half the MSR), padded (all features), bare (decision word only),
wrong-decision and distractor-only. Verbatim prompt texts ship in the
artifact.

\section{Qualitative Examples}
\label{app:qualitative}

Representative late-training rollouts, verbatim from the logged
\texttt{steps.jsonl} files. Scores are the logged per-rollout outputs of the
training proxy and the exact verifier.

\textbf{Hacked (sufficiency proxy, step 150; proxy $1.00$, true $0.55$,
causal precision $0.375$, 8 features cited).} \emph{``This tool call was
blocked by R4 (privilege-escalation) because the intern user with no
authentication and read-scope attempted to send an email containing both PII
and credentials to a public.web domain. Additionally, the request triggers R5
(session-risk) due to the intern's 2 prior violations and lack of
authentication.''} Five of the eight cited features are causally
irrelevant. Pinning everything makes the cited set trivially sufficient, so
the proxy scores it perfect.

\textbf{Faithful (exact reward, step 131; proxy $=$ true $=1.00$, precision
$1.0$, 4 features cited).} \emph{``This request was blocked by the
data-egress policy because the combination of basic authentication level with
access to a public web domain indicates a potential data egress.
Additionally, the intern-level user role accessing a public web domain with a
history of prior violations presents elevated risk for this action.''}

\textbf{Starved (outcome-only, step 105, the last step with non-zero group
variance; proxy $1.00$, true $0.67$, causal precision $0.50$, 4 features
cited).} \emph{``This tool call was blocked by the repeat-offender rule,
which prohibits requests containing personally identifiable information
(PII) with a prior violation on record. The combination of 1 documented
violation and PII presence satisfies the block criteria, preventing further
unauthorized data manipulation attempts. The call carries destructive intent
and lacks authentication, creating heightened security and compliance
risks.''} The outcome verifier awards this text, and every other rollout in
its group, the identical perfect score for containing the decision word. The
step therefore produces zero advantage, and the causally irrelevant intent
and authentication citations cost true precision that the training signal
never sees.

\textbf{Diluted (extraction noise $\varepsilon{=}.30$, step 150; proxy
$0.67$ computed on a post-flip citation set of 6, true $0.60$, causal
precision $0.43$, 7 features cited).} \emph{``This tool was blocked by the
default policy (R9) because no specific blocking policy matched, as the
intern user is sharing sensitive data externally, and the email.send scope
is being used. However, the intern's sensitive user role and prior
violations on record create additional risks, so the gateway defaults to
allowing sensitive external sharing. The intern's lack of authentication (no
auth\_level) combined with sensitive PII content prevents secure sharing, so
the sensitive-escalation policy (R4) blocks the action.''} This is the
rambling signature of the diluted phenotype. The text invokes two different
rules, contradicts its own decision narrative (``defaults to allowing''),
and cites seven features where a single minimal sufficient reason would do.
The noisy verifier meanwhile scored a citation set that differs from what
the text actually says.

\section{Full Curves and Transfer Breakdown}
\label{app:transfer}

\begin{figure}[H]
\centering
\includegraphics[width=\textwidth,height=0.78\textheight,keepaspectratio]{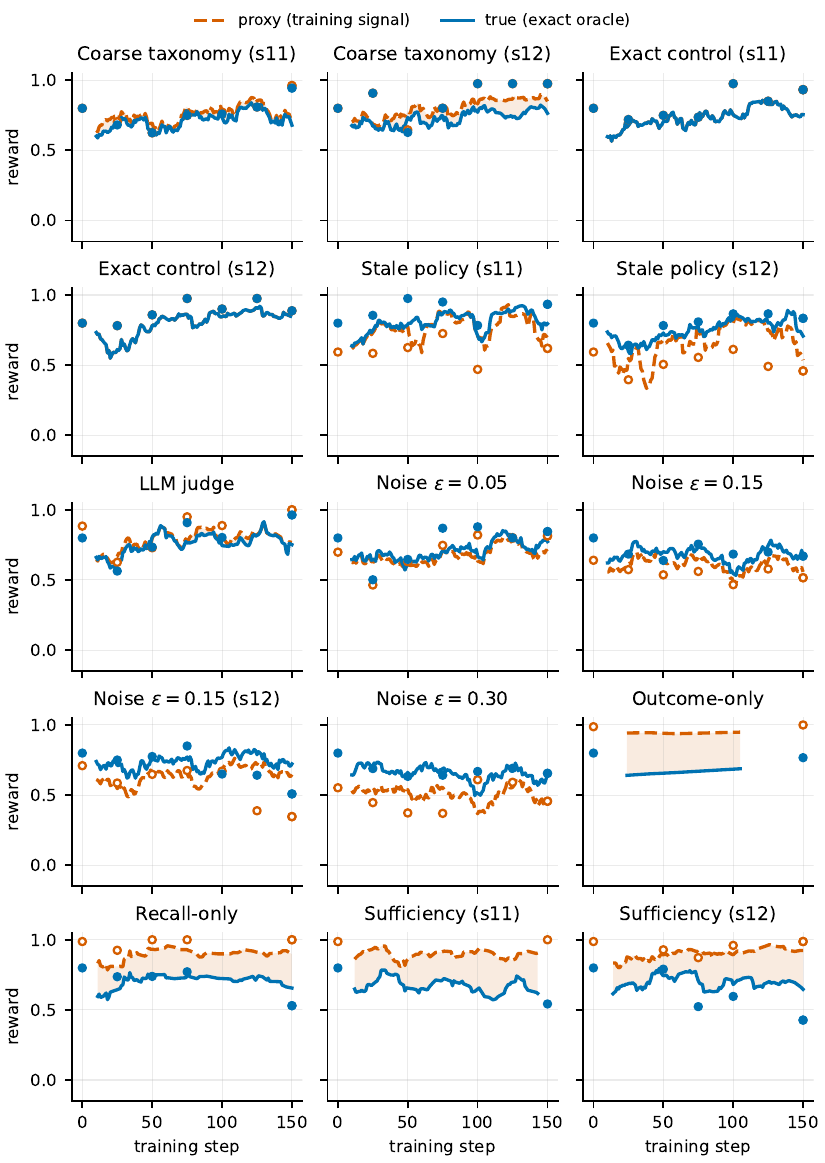}
\caption{Per-step proxy (dashed, vermillion) and exact true (solid, blue)
reward for all \NMlxRuns{} MLX runs. Circles are the 8-episode in-training
evals (terminal values in the text are re-scored at $n{=}\NTermEval$) and
the shaded band is the observed verifier gap where the proxy is
optimistic.}
\label{fig:curves_all}
\end{figure}

\paragraph{Full hacking curves.} Figure~\ref{fig:curves_all} shows the
per-step dual-reward trace for all \NMlxRuns{} MLX runs. The four
phenotypes of \S\ref{sec:anatomy} are visible as four distinct geometries.
The hacked cells (sufficiency, both seeds) open a band between proxy and
truth that widens from the first steps. The starved cell (outcome-only)
pins the proxy at ceiling, and the trace thins out because almost every step
is skipped for zero group variance. The diluted cells (noise) show the proxy
tracking truth from below, with a separation that grows in $\varepsilon$
while truth itself plateaus. The harmless cells are wrong about levels, with
the stale-policy proxy far below truth and the judge slightly above, while
truth stays at control level throughout.

\begin{figure}[H]
\centering
\includegraphics[width=0.47\textwidth]{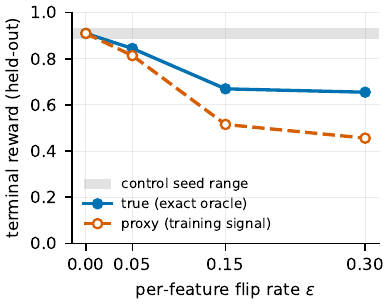}
\caption{\textbf{Noise dose-response.} Terminal held-out reward after 150
steps against the per-feature flip rate $\varepsilon$. The
$\varepsilon{=}0$ point is the mean of the two exact-control seeds and the
gray band their range. True reward (blue, filled) degrades gracefully and
saturates. The verifier's own report (vermillion, open) falls faster,
because per-feature noise is a pessimistic gap, so the trained policy ends
up better than its own training signal claims.}
\label{fig:dose}
\end{figure}

\begin{figure}[H]
\centering
\includegraphics[width=\textwidth,height=0.78\textheight,keepaspectratio]{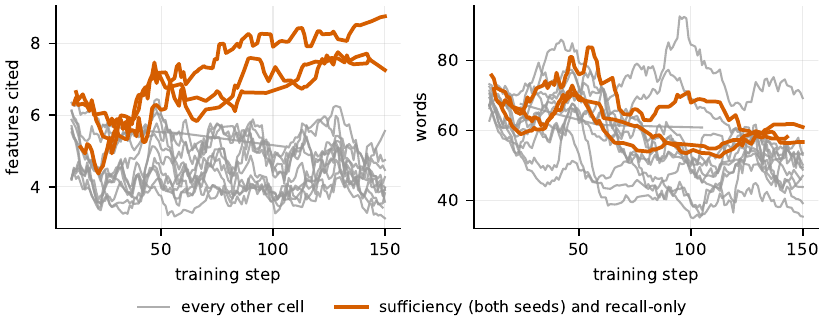}
\caption{\textbf{Behavioural signatures.} Rolling (window 10) per-step means
of features cited (left) and words emitted (right) for all \NMlxRuns{} MLX
runs, with the two sufficiency-proxy seeds highlighted (vermillion, thick).
They and the recall-only cell are the only runs whose citation counts
inflate beyond the initialisation band, the padding signature predicted
before training, while every other cell drifts flat or downward. Word counts do not separate the
hacked runs from the rest of the grid, so length-based monitoring
\citep{singhal2024length} alone would miss the hack.}
\label{fig:signatures}
\end{figure}

\paragraph{Per-policy transfer.} Table~\ref{tab:transfer} breaks the
aggregate transfer scores of \S\ref{sec:transfer} down by held-out policy.
No variant dominates across policies. PG-exact (seed 12) is strongest on
\texttt{opa-loadbalancer}, PG-exact (seed 11) on
\texttt{cedar-streaming-service}, and the untrained base model on
\texttt{cedar-tax-preparer} and \texttt{gk-psp-seccomp}. The hacked model's
table-topping aggregate decomposes exactly as its padding signature
predicts. Recall-saturated citation is competitive on every policy yet best
on only one (\texttt{gk-container-limits}), and its causal precision is the
best on none of the five. Aggregate $F_1$ rewards indiscriminate citation on
real policies just as it does on the synthetic grid.

\begin{table}[H]
\centering
\caption{Per-policy transfer on the five held-out real policies, as
causal-attribution $F_1$ against each policy's verified logical oracle
($n{=}16$ stratified episodes per policy, generation temperature $0.2$,
held-out LLM extractor). The bottom row is the episode-level mean over all
80 episodes and matches the aggregate numbers quoted in
\S\ref{sec:transfer}.}
\label{tab:transfer}
\resizebox{\textwidth}{!}{% AUTO-GENERATED by experiments/make_macros.py -- do not edit.
\begin{tabular}{lccccccc}
\toprule
Held-out policy & $n$ & Base & SFT & PG-exact (s11) & PG-exact (s12) & PG-suff.\ (hacked) & PG-single-gate \\
\midrule
\texttt{cedar-streaming-service} & 16 & 0.755 & 0.796 & 0.916 & 0.687 & 0.748 & 0.723 \\
\texttt{cedar-tax-preparer} & 16 & 0.823 & 0.640 & 0.531 & 0.740 & 0.768 & 0.710 \\
\texttt{gk-container-limits} & 16 & 0.623 & 0.663 & 0.690 & 0.644 & 0.767 & 0.734 \\
\texttt{gk-psp-seccomp} & 16 & 0.829 & 0.775 & 0.731 & 0.717 & 0.817 & 0.671 \\
\texttt{opa-loadbalancer} & 16 & 0.610 & 0.856 & 0.833 & 0.938 & 0.794 & 0.913 \\
\midrule
All episodes & 80 & 0.728 & 0.746 & 0.740 & 0.745 & 0.779 & 0.750 \\
\bottomrule
\end{tabular}
}
\end{table}

\paragraph{Statistical uncertainty.} Table~\ref{tab:transferci} attaches
episode-level uncertainty to the aggregate operating points of
\S\ref{sec:transfer}, as 95\% bootstrap confidence intervals (2{,}000
resamples over the 80 held-out episode scores, seed 0). All six $F_1$
intervals overlap pairwise, so no variant separates from any other on the
aggregate, which is the statistical form of the coverage null. On the
components, the one decisive difference is the hacked model's recall
saturation (\TransSuffRec{}, CI \TransSuffRecCI{}). A paired sign-flip
permutation test on per-episode recall against the base model gives
\PermSuffRecP{}. The PG-exact precision shift (\TransExactAPrec{} against
\TransBasePrec{} for the base) is directionally consistent across seeds but
not individually significant (\PermExactPrecP{} for seed 11 against base on
per-episode precision). We therefore describe it in \S\ref{sec:transfer} as
a shift of the operating point and not as a demonstrated gain.

\begin{table}[H]
\centering
\caption{Aggregate transfer scores with 95\% bootstrap confidence intervals
(2{,}000 resamples over the $n{=}80$ episode-level scores per variant).
Paired sign-flip permutation tests against the base model give hacked-model
recall \PermSuffRecP{} and PG-exact (seed 11) precision \PermExactPrecP.}
\label{tab:transferci}
% AUTO-GENERATED by experiments/make_macros.py -- do not edit.
\begin{tabular}{lccc}
\toprule
Model & Precision [95\% CI] & Recall [95\% CI] & $F_1$ [95\% CI] \\
\midrule
Base (no training) & 0.717 [0.63, 0.79] & 0.800 [0.72, 0.88] & 0.728 [0.65, 0.80] \\
SFT & 0.701 [0.63, 0.76] & 0.880 [0.82, 0.93] & 0.746 [0.69, 0.80] \\
PG-exact (s11) & 0.754 [0.68, 0.82] & 0.829 [0.77, 0.89] & 0.740 [0.68, 0.79] \\
PG-exact (s12) & 0.750 [0.67, 0.82] & 0.804 [0.73, 0.87] & 0.745 [0.68, 0.81] \\
PG-sufficiency (hacked) & 0.702 [0.64, 0.76] & 0.965 [0.94, 0.99] & 0.779 [0.73, 0.82] \\
PG-exact, single gate & 0.786 [0.71, 0.85] & 0.780 [0.71, 0.84] & 0.750 [0.69, 0.81] \\
\bottomrule
\end{tabular}

\end{table}

\section{Replication on an Independent Implementation}
\label{app:second-model}

This appendix gives the full record behind \S\ref{sec:replication}, whose
eight PyTorch configurations are listed in Table~\ref{tab:replication-all}.
Every terminal value is scored on the full held-out set ($n = \NTermEval$).
Seven sets of predictions were committed before their runs
(\texttt{docs/PREREG\_torch\_grid}, \texttt{\_headroom},
\texttt{\_matched\_recipe}, \texttt{\_quantisation}, \texttt{\_seed12},
\texttt{\_corpus} and \texttt{\_scale\_recall\_adapterfp32} in the
artifact), and each is scored below as it fell.
Table~\ref{tab:prereg-scoreboard} is the whole record on one page; the
paragraphs after it give the detail.

\begin{table}[H]
\centering
\caption{\textbf{Every replication prediction, scored.} One row per
pre-registered prediction, the quantity that decided it, and the verdict as
it fell. Nine held, six were refuted or not met, one held only within noise,
and the scale predictions could not be scored. Rows are grouped by the
paragraph below that gives the detail, and every number is a generated macro.
The corpus pre-registration (C1 to C5) is scored separately in
Appendix~\ref{app:corpus-grid}.}
\label{tab:prereg-scoreboard}
\scriptsize
\renewcommand{\arraystretch}{1.18}
\begin{tabular}{@{}l p{0.27\textwidth} p{0.335\textwidth} l@{}}
\toprule
\textbf{ID} & \textbf{What it required} & \textbf{What happened} & \textbf{Verdict} \\
\midrule
\multicolumn{4}{@{}l}{\textit{The first pre-registration}} \\
P5 & Sufficiency damage above stale-policy damage at every initialisation & Ordering holds in every configuration & held \\
unnamed & Diluted damage monotone in $\varepsilon$ & \RepMNoiseLoDam, \RepMNoiseMidDam, \RepMNoiseHiDam{}; the first two differ by less than their combined error & not met strictly, met within noise \\
\addlinespace
\multicolumn{4}{@{}l}{\textit{Headroom}} \\
P2 & Damage rises with policy competence as a dose-response & \RepTQLDamMean{} over two seeds, inside the standard-SFT band & refuted \\
P3 & More optimisation from a weak start produces the hack & \PhiThreeDam{} $\pm$ \PhiThreeDamSE{} at 300 steps, though inflation grew to \PhiThreeInfl & refuted \\
P4 & A stronger initialisation turns the null into a hack & Control \RepTPCtrl{} to \RepTPLCtrl{} turns \RepTPDam{} into \RepTPLDam{} $\pm$ \RepTPLDamSE & held \\
\addlinespace
\multicolumn{4}{@{}l}{\textit{Matched recipe}} \\
M1 & Matching the MLX adapter closes the magnitude gap & \RepTQDam{} at 1.3M parameters against \RepTQMDam{} at 7.3M & refuted \\
M3 & Harmless phenotypes reproduce under the matched adapter & Stale policy \RepMFamDam, coarse taxonomy \RepMCoarseDam, both inside the $0.05$ band & held \\
\addlinespace
\multicolumn{4}{@{}l}{\textit{Pessimistic cells and the sign asymmetry}} \\
P6 & Same sign asymmetry on the torch grid, with $\beta_+$ below its MLX value & \RepSlopeOpt{} against \RepSlopePess{} here, both below \SlopeOpt{} and \SlopePess{} on MLX & held \\
\addlinespace
\multicolumn{4}{@{}l}{\textit{Quantisation}} \\
Q1 & Low-bit weights produce a larger hack & NF4 gives \RepTQFourDam{} $\pm$ \RepTQFourDamSE, below the fp16 value & refuted \\
Q2 & Or the two are equal within the stated band & Falls just outside that band & not met \\
\addlinespace
\multicolumn{4}{@{}l}{\textit{Seed stability}} \\
unnamed & Each seed-12 damage within two combined standard errors of its seed-11 value & \SeedCellsWithin{} of \SeedCellsDone{} matched cells & held \\
\addlinespace
\multicolumn{4}{@{}l}{\textit{Recall-only verifier}} \\
R1 & Optimistic audit sign, selection error in $[0.10, 0.35]$ & Signed gap \RepRecallSigned, selection error \RepRecallSel & met \\
R2 & Hacked phenotype: inflation above 1.0, damage above 0.05 & Damage \RepRecallDam{} $\pm$ \RepRecallDamSE, inflation \RepRecallInfl & met \\
R3 & Adding the cell to the PyTorch-only fit keeps $\beta_+ > \beta_-$ & \RepSlopeOpt{} against \RepSlopePess & met \\
\addlinespace
\multicolumn{4}{@{}l}{\textit{Adapter precision}} \\
F1 & Damage above 0.15, which would name adapter precision as the cause & An fp32 adapter with no loss scaling lands \RepFpDelta{} from the fp16 value & not met \\
F2 & Damage within 0.04 of the fp16 value, which exonerates it & The same \RepFpDelta & met \\
\addlinespace
\multicolumn{4}{@{}l}{\textit{A larger model}} \\
L1 to L4 & Scale to Qwen3-8B in 4-bit NF4 with the matched adapter & SFT ran out of memory on a 16\,GB T4 twice & not scored \\
\bottomrule
\end{tabular}
\end{table}

\begin{table}[H]
\centering
\caption{\textbf{The sufficiency hack in every cell measured.} Damage is
the cell's own exact control minus its sufficiency cell at $n{=}\NTermEval$.
Citation inflation is the terminal cited-feature count of the sufficiency
cell minus its control's (Table~\ref{tab:training}'s Cited column is the
training-time first-to-last change instead). The optimistic gap (proxy
minus truth on the sufficiency cell) barely moves. The damage does. All
PyTorch cells fall in a narrow band, and the MLX grid is the outlier on both
damage and inflation. Figure~\ref{fig:replication}a plots the same rows.}
\label{tab:replication-all}
\resizebox{\textwidth}{!}{% AUTO-GENERATED -- do not edit.
\begin{tabular}{llcccc}
\toprule
Setup & Adapter & Control & Sufficiency damage & Cite infl. & Opt. gap \\
\midrule
MLX, Qwen3-4B, 4-bit (2 seeds) & 7.3M & 0.871 & +0.218 $\pm$ 0.018 & +4.29 & +0.310 \\
PyTorch fp16, Qwen3-4B & 1.3M & 0.758 & +0.084 $\pm$ 0.029 & +0.84 & +0.174 \\
PyTorch fp16, Qwen3-4B, long SFT, seed 11 & 1.3M & 0.791 & +0.061 $\pm$ 0.024 & +1.45 & +0.205 \\
PyTorch fp16, Qwen3-4B, long SFT, seed 12 & 1.3M & 0.819 & +0.096 $\pm$ 0.024 & +1.73 & +0.199 \\
PyTorch fp16, Qwen3-4B, matched adapter, seed 11 & 7.3M & 0.797 & +0.083 $\pm$ 0.028 & +2.15 & +0.193 \\
PyTorch fp16, Qwen3-4B, matched adapter, seed 12 & 7.3M & 0.812 & +0.093 $\pm$ 0.027 & +2.01 & +0.192 \\
PyTorch fp16, Phi-3.5-mini & 1.6M & 0.672 & +0.017 $\pm$ 0.039 & +0.99 & +0.217 \\
PyTorch fp16, Phi-3.5-mini, long SFT & 1.6M & 0.764 & +0.087 $\pm$ 0.031 & +1.54 & +0.236 \\
PyTorch fp16, Phi-3.5-mini, matched adapter & 7.3M & 0.702 & +0.071 $\pm$ 0.037 & +1.75 & +0.213 \\
PyTorch 4-bit NF4, Qwen3-4B, matched adapter & 7.3M & 0.780 & +0.034 $\pm$ 0.027 & +1.11 & +0.141 \\
PyTorch, Qwen3-4B, matched adapter held in fp32 & 7.3M & 0.826 & +0.076 $\pm$ 0.024 & +0.83 & +0.166 \\
\bottomrule
\end{tabular}
}
\end{table}

\begin{figure}[H]
\centering
\includegraphics[width=\textwidth,height=0.78\textheight,keepaspectratio]{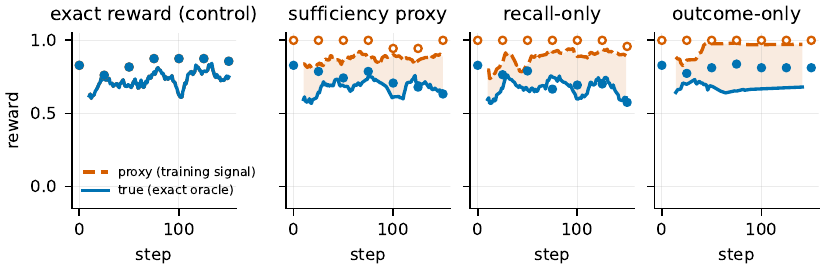}
\caption{\textbf{Per-step traces on the PyTorch matched-adapter grid.} The
same geometries as the MLX grid (Fig.~\ref{fig:curves_all}) at smaller
amplitude. The control closes the band, the sufficiency and recall-only
proxies open an optimistic band that truth falls away from, and the
outcome-only proxy pins to ceiling while the trace thins out as updates are
skipped.}
\label{fig:curves_torch}
\end{figure}

\paragraph{The first pre-registration.} \texttt{PREREG\_torch\_grid} was
written against the 8-episode fit ($\beta_+ = 3.29$) and the fixed-gate
audit of Table~\ref{tab:audit}. After every terminal value was re-scored on
\NTermEval{} episodes and the fit moved to the training-distribution audit
of Appendix~\ref{app:lawdata}, the slopes became \SlopeOpt{} and
\SlopePess{}, and Table~\ref{tab:replication} scores against that refit,
which predicts less damage and is the more conservative test. Against the original slopes
the predicted damages are 1.9 times larger and every optimistic cell misses
by more, so the conclusion does not depend on which slopes are used. Its criterion for the
diluted phenotype required damage monotone in $\varepsilon$. The replication
gives \RepMNoiseLoDam{}, \RepMNoiseMidDam{} and \RepMNoiseHiDam{}, which is
monotone only within one standard error (the first two differ by less than
their combined error of $0.04$), so that criterion is scored as not met
strictly and met within noise. The qualitative predictions held in every configuration: P5 (sufficiency
damage exceeds stale-policy damage at every initialisation, with the
stale-policy and coarse cells inside $0.05$ of their own control) and M3.

\begin{table}[H]
\centering
\caption{\textbf{The first replication grid (PyTorch, Qwen3-4B, q/v adapter)
against the pre-registered predictions.} Predicted damage is the MLX law's
slope times the cell's audited selection error and effective-update
fraction.}
\label{tab:replication}
% AUTO-GENERATED -- do not edit.
\begin{tabular}{lcccc}
\toprule
Cell & Audit sign & Skipped & Predicted & Observed \\
\midrule
Sufficiency proxy & optimistic & 27/150 & +0.311 & +0.084 $\pm$ 0.029 \\
Outcome-only & optimistic & 79/150 & +0.389 & +0.033 $\pm$ 0.028 \\
Stale policy & pessimistic & 0/150 & +0.051 & -0.004 $\pm$ 0.028 \\
Coarse taxonomy & optimistic & 0/150 & +0.065 & +0.040 $\pm$ 0.031 \\
\bottomrule
\end{tabular}

\end{table}

\paragraph{Headroom (\texttt{PREREG\_headroom}).} P2 predicted that damage
would rise with the policy's competence in a dose-response fashion. It did
not. A longer SFT on Qwen gave damages of \RepTQLDamA{} and \RepTQLDamB{} at
two seeds (mean \RepTQLDamMean), inside the band of the standard SFT. P3
continued the Phi cells from 150 to 300 steps and found damage of
\PhiThreeDam{} $\pm$ \PhiThreeDamSE{} while citation inflation kept growing
to \PhiThreeInfl{}. More optimisation from a weak start does not produce
the hack. P4 held. Raising Phi's initialisation from a control of
\RepTPCtrl{} to \RepTPLCtrl{} turned a null (\RepTPDam) into a hack of
\RepTPLDam{} $\pm$ \RepTPLDamSE{} with nothing else changed. Competence
therefore gates whether the hack appears, as a threshold rather than a
dose.

\paragraph{Matched recipe (\texttt{PREREG\_matched\_recipe}).} M1 predicted
that matching the adapter to the MLX one (every projection, 7.3M
parameters) would close the magnitude gap. It was refuted. Qwen gives
\RepTQDam{} with the query/value adapter and \RepTQMDam{} with the matched
one. M3, that the harmless phenotypes reproduce under the matched adapter,
held (stale policy \RepMFamDam{}, coarse taxonomy \RepMCoarseDam{}).

\paragraph{Pessimistic cells and the sign asymmetry (P6).} Extraction noise
at $\varepsilon = .05$, $.15$ and $.30$ costs \RepMNoiseLoDam{},
\RepMNoiseMidDam{} and \RepMNoiseHiDam{} against the matched control at
signed gaps of \RepMNoiseLoGap{}, \RepMNoiseMidGap{} and \RepMNoiseHiGap.
These pessimistic cells do damage, which is what makes a fit on this grid
alone informative. The fit is given
in \S\ref{sec:replication} and Fig.~\ref{fig:law}b.

\paragraph{Quantisation (\texttt{PREREG\_quantisation}).} On 4-bit NF4
weights the sufficiency cell does \RepTQFourDam{} $\pm$ \RepTQFourDamSE{}
of damage at an optimistic gap of \RepTQFourGap{}. That is below the fp16
value rather than above it, so low-bit weights do not produce a larger hack
and Q1 fails. It also falls just outside the band we set for calling the two
equal (Q2), so under our own rule we attribute the MLX magnitude to neither
and leave it open. The stale-policy and coarse cells stay inside the
harmless band once more (\RepMFourFamDam{} and \RepMFourCoarseDam).

\paragraph{Seed stability (\texttt{PREREG\_seed12}).} Every matched-adapter
cell was re-run at seed 12, with the criterion that each seed-12 damage
falls within two combined standard errors of its seed-11 value.
Table~\ref{tab:seed12} reports the pairs. \SeedCellsWithin{} of
\SeedCellsDone{} cells meet the criterion, and the seed-12 control is
\SeedBCtrl.

\begin{table}[H]
\centering
\caption{\textbf{Seed replication of the PyTorch matched-adapter grid.}
Damage against each seed's own exact control, $n{=}\NTermEval$.}
\label{tab:seed12}
% AUTO-GENERATED -- do not edit.
\begin{tabular}{lccccc}
\toprule
Cell & Damage, seed 11 & Damage, seed 12 & $|\Delta|$ & $2\,\mathrm{SE}$ & Within \\
\midrule
Suff & +0.083 & +0.093 & 0.011 & 0.078 & yes \\
Outcome & +0.019 & +0.094 & 0.075 & 0.082 & yes \\
Fam & -0.006 & -0.013 & 0.008 & 0.077 & yes \\
Coarse & +0.010 & +0.019 & 0.009 & 0.073 & yes \\
NoiseLo & +0.024 & +0.001 & 0.023 & 0.079 & yes \\
NoiseMid & +0.018 & +0.012 & 0.007 & 0.074 & yes \\
NoiseHi & +0.071 & +0.068 & 0.003 & 0.080 & yes \\
\bottomrule
\end{tabular}

\end{table}

\paragraph{Recall-only verifier (\texttt{PREREG\_scale\_recall\_adapterfp32},
part B).} R1 required an optimistic audit sign with selection error between
0.10 and 0.35, and the family-distribution audit gives \RepRecallSigned{}
and \RepRecallSel{}, so R1 is met. R2 required the hacked phenotype with
citation inflation above 1.0 and damage above 0.05 against the matched
control. The cell ends at \RepRecallTrue{} (damage \RepRecallDam{} $\pm$
\RepRecallDamSE{}) citing \RepRecallCited{} features (inflation
\RepRecallInfl{}) while the proxy reports \RepRecallProxy{}, with
\RepRecallSkips{} skipped updates, so R2 is met. R3 required that adding
the cell to the PyTorch-only fit keep $\beta_+ > \beta_-$, and it does
(\RepSlopeOpt{} against \RepSlopePess). Figure~\ref{fig:curves_torch}
shows the trace.

\paragraph{Adapter precision (part C).} MLX trains an fp32 adapter over a
frozen 4-bit base with no loss scaling, whereas the PyTorch port trains an
fp16 adapter under a gradient scaler. The testable variant on a T4 holds
the LoRA parameters in fp32 with the scaler disabled and everything else
matched. F1 (damage above 0.15, which would name adapter precision as the
cause) and F2 (damage within 0.04 of the fp16 value, which exonerates it)
were the two outcomes. The cell gives a control of \RepTQFpCtrl{} and
damage of \RepTQFpDam{} $\pm$ \RepTQFpDamSE{}, which is \RepFpDelta{} from
the fp16 value. Verdict, \RepFpVerdict. Adapter capacity, base quantisation
and adapter precision are now each excluded as the explanation of the MLX
magnitude.

\paragraph{A larger model (part A).} We pre-registered Qwen3-8B in 4-bit
NF4 with the matched adapter. The SFT stage ran out of memory on the
16\,GB T4 twice, the second time at batch 1, sequence length 1024 and
gradient accumulation 2, so L1 to L4 could not be scored and the larger
model is recorded as infeasible on this hardware rather than as a result in
either direction.

\paragraph{What the replication changes.} The gap
law's constants are specific to the MLX grid. Its sign, its ordering and its
phenotype predictions are not, and the audit quantity it rests on is
invariant across every configuration we ran. The sufficiency hack replicates
in existence and signature but not in magnitude. The strong form of the
claim, that a given verifier gap causes a given amount of damage, is not
supported and is not made.

\section{Training on the Real-Policy Corpus}
\label{app:corpus-grid}

Every other grid trains on the synthetic GlassGate family. Here the
\NCorpusPolicies{} ported Cedar, OPA and Gatekeeper policies of
Appendix~\ref{app:corpus} are the training domain. The
policy gates are exact deny-overrides oracles over 36 to 3,456 contexts
each, the extractor is a schema-generic deterministic mention detector over
each policy's own feature names (it recovers the rendered probe suite's
cited sets with \CorpProbeMisses{} misses in \CorpProbeTotal{} probes), and
the training prompts render each episode's context and policy. Nine
policies train, five are held out, and all terminal values are scored on
the 80 held-out episodes. Predictions were committed before the run
(\texttt{PREREG\_corpus.md}, with an amendment before the base-model run).

\paragraph{The audit transfers before any training.} On the corpus probe
suite the outcome-only verifier's selection error is \CorpSelErrOutcome{} (a
coin flip, as on GlassGate) with signed gap \CorpSignedOutcome. The
sufficiency proxy is optimistic (\CorpSignedSuff, selection error
\CorpSelErrSuff). Extraction noise (\CorpSignedNoiseMid) and a stale policy
(\CorpSignedDrop) are pessimistic. None of these numbers depended on a
policy model.

\paragraph{Two attempts that could not train.} The first attempt
used SFT targets templated from each episode's canonical minimal reason.
That collapsed the policy to one deterministic text per prompt (144 rows, 60
distinct completions), so every group had zero reward variance. The second
attempt varied the targets, and the policy still skipped 124 of 150 steps
because the SFT had already solved the task, with mean group reward rising
from 0.978 to 0.999 and the held-out eval flat across all checkpoints. Real
minimal reasons are one or two features, so an SFT on exactly those answers
leaves no headroom. We amended the pre-registration to train from the base model with no corpus
SFT, and added a trainability gate: C5 requires a step-1 group reward below
0.8 and fewer than 60 skipped control steps. The original four, unchanged,
cover the audit signs (C1), the four phenotypes with their
citation-inflation and saturation thresholds (C2), a control that does not
lose true reward against the SFT initialisation (C3), and the
optimistic-steeper sign asymmetry (C4).

\paragraph{Training from the base model.} The untrained base scores
\CorpBase{} $\pm$ \CorpBaseSE{} on the held-out episodes. From that start
the exact control reaches \CorpCtrl{} $\pm$ \CorpCtrlSE{} (citing
\CorpCtrlCited{} features) while skipping \CorpCtrlSkips{} of 150 updates,
and the four degraded verifiers give the damages in
Table~\ref{tab:corpus-grid}.
\begin{figure}[H]
\centering
\includegraphics[width=0.62\textwidth]{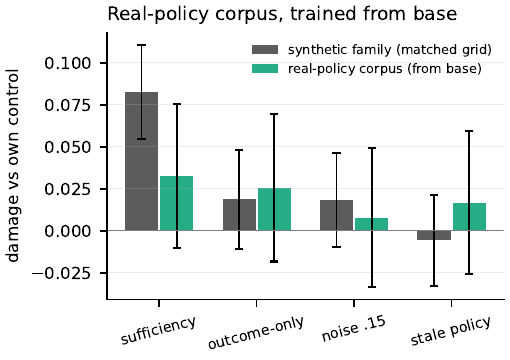}
\caption{\textbf{The corpus grid against the synthetic family.} Damage of
each degraded verifier against its own exact control, on the real-policy
corpus trained from the base model (green) and on the synthetic family under
the matched adapter (grey). The sufficiency cell is largest in both, but the
corpus cells are starved, skipping 116, 149, 8 and 96 of 150 updates from
left to right, so none of their damages is distinguishable from zero.}
\label{fig:corpusgrid}
\end{figure}

\begin{table}[H]
\centering
\caption{\textbf{Degraded-verifier grid trained on real public policies
from the base model} (150 PG steps, matched adapter, seed 11, $n{=}80$
held-out episodes).}
\label{tab:corpus-grid}
% AUTO-GENERATED -- do not edit.
\begin{tabular}{lcccc}
\toprule
Training verifier & True$_{150}$ & Proxy$_{150}$ & Damage & Cite infl. \\
\midrule
Sufficiency proxy & 0.752 & 0.929 & +0.033 $\pm$ 0.043 & +0.21 \\
Outcome-only & 0.759 & 1.000 & +0.026 $\pm$ 0.044 & -0.04 \\
Extraction noise $\varepsilon{=}.15$ & 0.777 & 0.706 & +0.008 $\pm$ 0.041 & +0.05 \\
Stale policy & 0.768 & 0.611 & +0.017 $\pm$ 0.043 & -0.06 \\
\bottomrule
\end{tabular}

\end{table}
The sufficiency cell does \CorpSuffDam{} $\pm$ \CorpSuffDamSE{} of damage
with citation inflation \CorpSuffInfl{} and \CorpSuffSkips{} skipped
updates. The outcome proxy saturates at \CorpOutcomeProxy{} and skips
\CorpOutcomeSkips{} updates for \CorpOutcomeDam{} of damage. Noise costs
\CorpNoiseMidDam{} and the stale policy \CorpDropDam{}. Scored against
\texttt{PREREG\_corpus.md}, the record reads \CorpScore. Because C5 is not
met, the pre-registration's own rule applies. The corpus contributes its
audit, which transferred, and a descriptive grid in which only the identity
of the top cell (sufficiency) matches the synthetic family and every damage
is within about one standard error of zero. It does not contribute a second measurement of damage. The control's gain over the base
(\CorpCtrlGain{} $\pm$ \CorpCtrlGainSE) is likewise directional only.

\end{document}